\documentclass[ijds,nonblindrev]{informs-ijds}

\OneAndAHalfSpacedXI

\usepackage{natbib}
 \bibpunct[, ]{(}{)}{,}{a}{}{,}%
 \def\bibfont{\small}%

\usepackage{hyperref}
\hypersetup{
    colorlinks=true,
    citecolor=blue,
    linkcolor=blue,
    filecolor=blue,      
    pdftitle={Overleaf Example},
    pdfpagemode=FullScreen,
    }
    
\usepackage{setspace}
\usepackage{multirow}
\usepackage{adjustbox}
\usepackage{makecell}
\usepackage{pifont}
\usepackage[ruled,vlined]{algorithm2e}
\usepackage{color}
\newcommand{\xmark}{\ding{55}}%
\newcommand{\fm}{f_m}
\newcommand{\Fm}{\mathbf{f}_{m}}
\newcommand{\Fmdash}{\mathbf{f}_{m'}}
\newcommand{\N}{\mathcal{N}}
\newcommand{\bu}{\mathbf{u}}
\newcommand{\Z}{\mathbf{Z}}
\newcommand{\z}{\mathbf{z}}
\newcommand{\X}{\mathbf{X}}
\newcommand{\x}{\mathbf{x}}
\newcommand{\Y}{\mathbf{y}}
\newcommand{\y}{\mathbf{y}}
\newcommand{\Xu}{\mathbf{X}^u}
\newcommand{\Xl}{\mathbf{X}^l}
\newcommand{\F}{\mathbf{f}}
\newcommand{\zero}{\mathbf{0}}
\newcommand{\K}{\mathbf{K}}
\newcommand{\KFF}{\mathbf{K_{f,f}}}

\newcommand{\KFu}{\K_{\mathbf{f},\mathbf{u}}}
\newcommand{\KuF}{\K_{\mathbf{u},\mathbf{f}}}
\newcommand{\KuFm}{\K_{\mathbf{u},\mathbf{f}_m}}
\newcommand{\KFmu}{\K_{\mathbf{f}_m,\mathbf{u}}}
\newcommand{\Kuu}{\mathbf{K_{\mathbf{u},\mathbf{u}}}}
\newcommand{\Kuuinv}{\mathbf{K}^{-1}_{\mathbf{u},\mathbf{u}}}
\newcommand{\Kfmfm}{\mathbf{K_{\mathbf{f}_{m},\mathbf{f}_{m}}}}
\newcommand{\bSigma}{\mathbf{\Sigma}}
\newcommand{\bPi}{\mathbf{\Pi}}
\newcommand{\bpi}{\boldsymbol{\pi}}
\newcommand{\balpha}{\boldsymbol{\alpha}_0}
\newcommand{\bbalpha}{\boldsymbol{\alpha}}
\newcommand{\B}{\mathbf{B}}
\newcommand{\bdiag}{\text{bdiag}}
\newcommand{\I}{\mathbf{I}}
\newcommand{\G}{\mathbf{G}}
\newcommand{\w}{\mathbf{w}}
\newcommand{\W}{\mathbf{W}}
\newcommand{\bL}{\mathbf{L}}
\newcommand{\KL}{\text{KL}}
\newcommand{\D}{\mathbf{D}}

\newcommand{\bA}{\mathbf{A}}
\newcommand{\bP}{\mathbf{P}}
\newcommand{\bQ}{\mathbf{Q}}
\newcommand{\bH}{\mathbf{H}}
\newcommand{\bC}{\mathbf{C}}
\newcommand{\bJ}{\mathbf{J}}

\newcommand{\bS}{\mathbf{S}}
\newcommand{\bG}{\mathbf{G}}
\newcommand{\bO}{\mathbf{O}}
\newcommand{\bR}{\mathbf{R}}

\newcommand{\ba}{\mathbf{a}}
\newcommand{\bv}{\mathbf{v}}
\newcommand{\Ainv}{\mathbf{A}^{-1}}
\newcommand{\Binv}{\mathbf{B}^{-1}}

\newcommand{\btheta}{\boldsymbol{\theta}}
\newcommand{\bsigma}{\boldsymbol{\sigma}}
\newcommand{\Lcvb}{\mathcal{L}_{\text{cVB}}}
\newcommand{\Lsvb}{\mathcal{L}_{\text{sVB}}}

\newcommand{\bmu}{\boldsymbol{\mu}}

\newcommand{\mvec}{\textbf{:}}
\newcommand{\Lcvbone}{\mathcal{L}^{(1)}_{\text{cVB}}}
\newcommand{\Lsvbone}{\mathcal{L}^{(1)}_{\text{sVB}}}
\newcommand{\Lsvbtwo}{\mathcal{L}^{(2)}_{\text{sVB}}}

\usepackage{amsmath}

\TheoremsNumberedThrough     
\ECRepeatTheorems

\EquationsNumberedThrough    

\begin{document}


\RUNAUTHOR{Chung, Kontar, and Wu}

\RUNTITLE{Weakly-supervised Multi-output Regression via Correlated Gaussian Processes}

\TITLE{Weakly-supervised Multi-output Regression via Correlated Gaussian Processes
}

\ARTICLEAUTHORS{%
\AUTHOR{Seokhyun Chung}
\AFF{Department of Industrial \& Operations Engineering, University of Michigan, Ann Arbor, MI 48109, \EMAIL{seokhc@umich.edu}} 
\AUTHOR{Raed Al Kontar}
\AFF{Department of Industrial \& Operations Engineering, University of Michigan, Ann Arbor, MI 48109, \EMAIL{alkontar@umich.edu}}
\AUTHOR{Zhenke Wu}
\AFF{Department of Biostatistics, University of Michigan, Ann Arbor, MI 48109, \EMAIL{zhenkewu@umich.edu}}
} 

\ABSTRACT{%
Multi-output regression seeks to borrow strength and leverage commonalities across different but related outputs in order to enhance learning and prediction accuracy. A fundamental assumption is that the output/group membership labels for all observations are known. This assumption is often violated in real applications.  For instance, in healthcare datasets, sensitive attributes such as ethnicity are often missing or unreported. To this end, we introduce a weakly-supervised multi-output model based on dependent Gaussian processes. Our approach is able to leverage data without complete group labels or possibly only prior belief on group memberships to enhance accuracy across all outputs. Through intensive simulations and case studies on an Insulin, Testosterone and Bodyfat dataset, we show that our model excels in multi-output settings with missing labels, while being competitive in traditional fully labeled settings. We end by highlighting the possible use of our approach in fair inference and sequential decision-making. 
}%




\maketitle

%


\section{Introduction}\label{sec::introduction}

Multi-output regression aims to estimate multiple unknown or latent functions for observations arising from a known number of outputs/groups. In each group, the inferential target is a function that relates a possibly multivariate vector of covariates to a univariate outcome. Relative to a separate regression for each group, multi-output regression is statistically appealing as it can borrow strength by exploiting potential between-group similarities resulting in improved learning and prediction accuracy. This is related to the general principle of hierarchical Bayesian models or multilevel models, where separate estimates may vary greatly in precision necessitating statistical borrowing of information between groups of varying sizes. For instance, healthcare datasets that aim to fit regression curves across different ethnicities may benefit from an integrative multi-output analysis so that minority groups can leverage knowledge from groups with more data.

Indeed, in recent years multi-output regression has seen great success in statistics, machine learning and many applications, specifically using Gaussian processes (GP) \citep[e.g.,][]{gu2016parallel, lee2019bayesian, diana2020hierarchical}. One primary technical convenience afforded by GP is that correlated outputs can be expressed as a realization from a single GP, where commonalities are modeled through inducing cross-correlations between outputs \citep[e.g.,][]{ver1998constructing}. Besides that, the uncertainty quantification capability of GPs and their ability to handle flexible model priors have rendered them especially useful in decision-making applications amongst others such as Bayesian optimization \citep{frazier2018tutorial}, reinforcement learning and bandits \citep{srinivas2009gaussian}, experimental design \citep{gramacy2015local}, computer experiments \citep{ba2012composite}, calibration and reliability \citep{kontar2018nonparametric}. In this paper, we refer to this class of models based on GP for estimating multiple outputs as ``multi-output GP'' (MGP).

\begin{figure}[h!]
	\centering
	\caption{Examples of fully labeled (left) and partially labeled (right) settings for multi-output regression.}
	\vspace{0.2in}
	\includegraphics[width=0.8\textwidth]{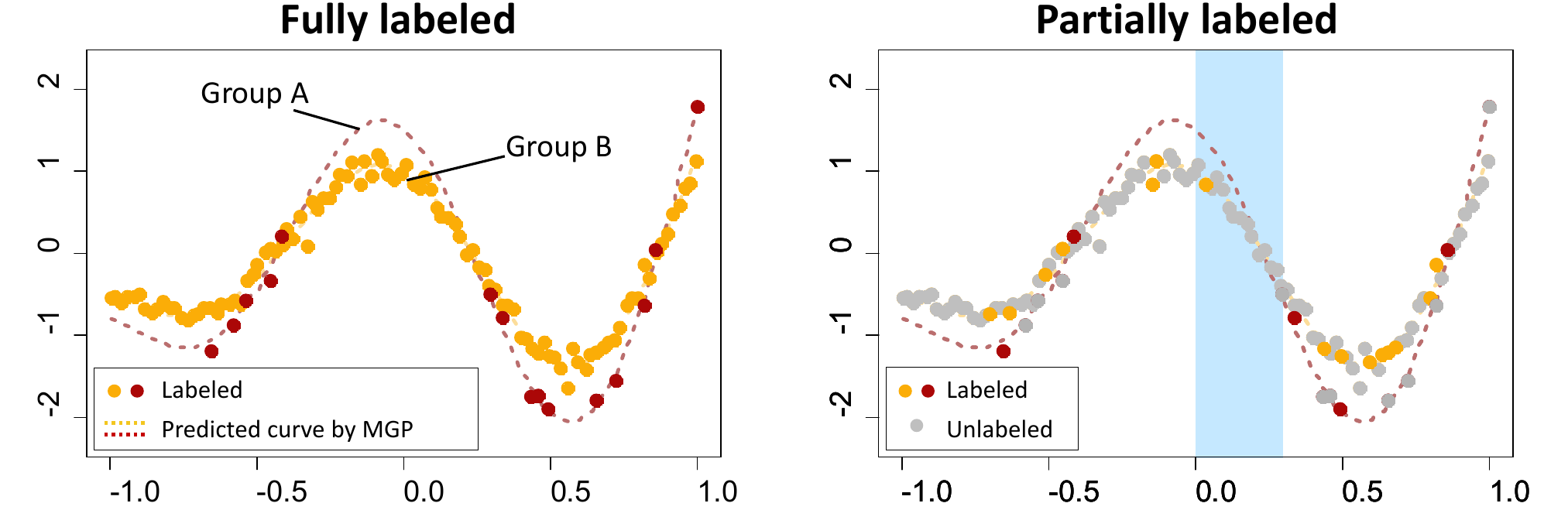}
	\label{fig:probDesc}
\end{figure}

{One fundamental assumption in existing MGPs is the availability of data with complete group membership labels. In many applications, group labels are often hard or expensive to obtain. A simple example is that participants might choose to keep their ethnicity or gender data private. As a result, multi-output regression that aims to borrow strength across different ethnicities to infer latent functions, becomes a rather challenging task. Hence, there exists an acute need for methods that can handle data without complete group labels or possibly only prior belief on group memberships. Such a class of regression problems where labels are missing (i.e. unsupervised) or only prior belief exists has recently been referred to as weakly-supervised learning \citep[e.g.,][]{grandvalet2005semi,singh2009unlabeled,rohrbach2013transfer, ng2018bayesian}.  In the context of multi-output regression, the weakly-supervised setting is first illustrated in a hypothetical scenario in Fig. \ref{fig:probDesc}. }

To illustrate, Fig. \ref{fig:probDesc} compares the simulated settings for fully labeled (left) and partially labeled (right) multi-output regression problems. The underlying curves for the two groups, A and B, although different, exhibit similar shapes over the range of a single continuous covariate. However, group A has sparser observations along the curve to inform the underlying truth compared to group B. The classical multi-output regression problem corresponds to the fully labeled case (left), where all the observations with complete group labels are used to estimate group-specific regression curves while accounting for between-group similarities. In our setting with incomplete group labels (right), some observations do not have observed group labels (grey dots), hindering the direct application of classical multi-output regression in data analysis. A simple solution is to remove observations with incomplete group labels. This however is suboptimal and may reduce estimation and prediction accuracy. For example, in the highlighted vertical band around zero, there are observations that do not have group labels but are close to observations from group B. An ideal approach should leverage these observations without group labels,  predict their group membership of B with high confidence and simultaneously use the memberships in curve estimation for group B.

This paper tackles the exact problem above. Specifically, we first propose a weakly-supervised regression framework based on correlated GPs that is able to learn missing group labels and simultaneously estimate conditional means and variances for all groups by exploiting their similarities. Our approach assigns a group membership indicator following a multinomial distribution for all observations. For observations with unknown group labels, we specify a Dirichlet prior for the multinomial probabilities with hyperparameters to be optimized. The Dirichlet prior acts as a regularizer that controls group assignment ``acuteness'', which upon optimization over a suitable objective function reduces predictive variance at each covariate value. For observations with observed group labels, the model can use prior belief on possibly noisy label memberships. We then propose a variational inference (VI) framework to estimate the intractable posterior. Our VI approach features a tractable lower bound on the intractable marginal likelihood and enjoys desirable computational and practical properties. To the best of our knowledge, this is the first study that addresses the weakly-supervised setting in MGP models. 

We conduct extensive simulations and case studies to demonstrate the operating characteristics of our modeling framework and compare it to other common alternatives. Our case studies feature three healthcare datasets (\textit{insulin}, \textit{testosterone}, and \textit{bodyfat}) and highlight the ability of our proposed model to leverage unlabeled data and improve performance over a range of weakly-supervised settings including semi-supervision and prior belief on group memberships. The insulin dataset collects insulin and C-peptide levels in the blood of individuals. Estimating latent curves relating the insulin to the C-peptide level is important as C-peptide is often used as an indicator of how much insulin is being produced \citep{leighton2017practical}. Such a relationship needs to be estimated within two groups: people taking insulin or not. A multi-output regression is able to achieve this task through estimating latent curves with both shared and unique features across the groups. However, a significant portion of the participants does not report whether they are taking insulin or not. This in turn necessitates an approach that takes unlabeled group membership into consideration. A similar data settings with missing group labels is also observed within the testosterone and bodyfat datasets. All the three case studies showcase the benefits of our proposed model and shed light on its ability to leverage missing data and borrow strength across groups for improved prediction and uncertainty quantification.

Below we summarize our contributions.

\begin{itemize}

\item We introduce a weakly-supervised regression framework based on correlated GPs that efficiently infers missing group membership labels and estimates conditional means and variances for multiple correlated groups. The proposed model can leverage unlabeled observations, resulting in more accurate predictions. To the best of our knowledge, this is the first work that tackles multi-output regression with missing group labels.
    
    \item We derive a modified variational bound for model inference that brings several practical advantages: (i) Ease of modeling whereby any MGP construction can be readily plugged into the lower bound (Sec. \ref{sec::ease}), (ii) Interpretability which might help design an optimization algorithm specialized to the lower bound (Sec. \ref{sec::interprete}), (iii) Amenability of the lower bound to mini-batch stochastic optimization (Sec. \ref{sec::scalable}), and (iv) Controllable labeling acuteness which enables the model to have better flexibility (Sec. \ref{sec::dirichlet}).
    
    \item We apply the proposed model and inferential algorithm to three real-world healthcare datasets that include unlabeled data. We demonstrate that our model can leverage unlabeled data and potential between-group correlation to remarkably improve predictive accuracy.

\end{itemize}

The rest of the paper is organized as follows. To contextualize our work, we first provide a brief review of GPs and MGPs in Sec. \ref{sec::background}. Sec.  \ref{sec::weak_supervision} formulates the proposed probabilistic model with weak-supervision designed for data with incomplete group labels. Then Sec. \ref{sec::inference} introduces a variational inference framework to estimate the intractable posterior and overcome the computational burden of serial sampling schemes. Sec.  \ref{sec::prediction} derives the predictive distribution for all outputs. Sec. \ref{sec::related_work} reviews related literature and compares the developed model to an unsupervised model and a fully-supervised model.
The utility of our model is highlighted using both simulated examples and real-world healthcare data in Sec. \ref{sec:experiment}. The paper concludes in Sec. \ref{sec::conclusion} with a discussion. We note that an additional real-world study on a bodyfat dataset and detailed derivations for the presented methodology are deferred to the Appendix.

\section{Background}
\label{sec::background}

In this section, we briefly review GP and MGP regression.

\subsection{Gaussian Processes Regression}
A GP is a stochastic process where the collection of any finite number of random variables is assumed to follow a multivariate Gaussian distribution \citep{Rasmussen06gaussianprocesses}. When a regression problem is modeled using GP, we refer it to as the GP regression. Specifically, consider that we are given $N_0$ observations composed of (multivariate) input $\X = [\x_n]^\top_{n=1,\dots,N_0} \in \mathbb{R}^{N_0\times d}$ and output $\y = [y_n]^\top_{n=1,\dots,N_0} \in \mathbb{R}^{N_0\times 1}$. We assume the output $\y$ is a realization of a true latent value corrupted by Gaussian white noise and expressed as $ y_n = f(\x_n) + \epsilon_n$ for $n\in \{1,\dots,N_0\},$ where $f(\cdot)$ denotes the true latent function and $\epsilon_n \sim \N(0,\sigma^2)$ denotes i.i.d. Gaussian noise with variance $\sigma^2$. That is, $\y \sim \N(\mathbf{f}_{\sf GP}, \sigma^2 \I)$ where $\mathbf{f}_{\sf GP} = [f(\x_n)]^\top_{n=1,\dots,N_0}$ and $\I$ is the identity matrix. In GP regression, we place a multivariate Gaussian prior on the latent values $\mathbf{f}_{\sf GP}$ expressed as $\mathbf{f}_{\sf GP} \sim \N(\mathbf{0}, \K_{\sf GP})$, where $\K_{\sf GP}$ indicates a $N_0\times N_0$ covariance formed by elements $k(\x_n,\x_{n'};{\btheta})=\text{cov}(f(\x_n),f( \x_{n'}))$ for $n,n' \in \{1,\cdots,N_0\}$ with a kernel $k(\cdot,\cdot)$ governed by the hyperparameter set $\btheta$. 

An advantageous feature of GP regression is that a closed form of the posterior predictive distribution is available. Let $\x^*$  denote a new input and $\mathbf{k}^{*} = [k(\x_n, \x^*)]^\top_{n=1,\dots,N_0}$, which is the corresponding $N_0\times 1$ covariance vector with $\X$. The predictive distribution for the corresponding output $\y^*$ is then derived by \begin{align*}
    p(\y^*|\X, \y, \x^*;\btheta,\sigma) &= \int p(\y^* | \x^*, \F_{\sf GP};\sigma) p(\F_{\sf GP}|\X, \y; \btheta) d\F_{\sf GP} \\ 
    &=  \N(\mathbf{k}^{*\top}(\K_{\sf GP}+\sigma^2\I)^{-1}\y, k(\x^*,\x^*) - \mathbf{k}^{*\top}(\K_{\sf GP}+\sigma^2\I)^{-1}\mathbf{k}^{*}+\sigma^2).
\end{align*}

The hyperparameters $\btheta$ and $\sigma$ are often estimated by maximizing the marginal likelihood expressed as
\begin{align*}
\log p(\y|\X ; \btheta,\sigma) = \log \int p(\y|\F_{\sf GP}; \sigma) p(\F_{\sf GP}| \X ; \btheta)d\F_{\sf GP} = \log \N(\y| \mathbf{0}, \K_{\sf GP}+\sigma^2\I).  
\end{align*}
One should note that maximizing the marginal log-likelihood requires the inversion of the $N_0 \times N_0$ covariance matrix, which induces time complexity $\mathcal{O}(N_0^3)$.

\subsection{Multi-output Gaussian Processes for Multi-output Regression}
\label{sec::MGP}

The main idea in multi-output regression analysis is establishing a statistical model which induces correlation across multiple outputs. Given the characteristic of GP that features correlations in the observations, the inter-output correlations can be naturally integrated into a GP with a large covariance matrix composed of block matrices for cross-covariance. Consider that we have $M$ outputs with inputs $\X_m=[\x_{mn}]^\top_{n=1,\dots,N_m}$ where the output from $m$-th group (or source) is 
composed of $N_m$ observations. The entire output is formed by concatenating the outputs as $\y = [\y_1^\top,\dots,\y_M^\top]^\top$ with the output from $m$-th group $\y_m = [y_{mn}]^\top_{n=1,\dots,N_m}$ for $m \in \{1,\dots,M\}$.

Like a standard GP regression, we assume the vector of outputs obtained at distinct input values follows $y_{mn} = f_m(\x_{mn})+\epsilon_{mn}$, for observation $n=1, \ldots, N_m$, where $f_m(\cdot)$ and $\epsilon_{mn} \sim \N(0,\sigma_m^2)$ denote the $m$-th unknown regression function and an independent additive Gaussian noise, respectively. For compact notation, let $\F=[\F_1^\top, \ldots, \F^\top_M]^\top$ collectively denote the latent values, where the column vector $\F_m = f_m(\X_m) = [f_m(\x_{mn})]^\top_{n=1,\dots, N_m}$ collects the latent function values from $f_m$ at each input point in group $m \in \{1,\dots,M\}$. Here, an MGP considers a $N_m \times N_{m'}$ cross-covariance matrix $\K_{\Fm,\Fmdash}$ for the group $m$ and $m'$ with elements $k_{f_m,f_{m'}}(\X_m, \X_{m'}; \btheta_{f_m,f_{m'}}) = \text{cov}(f_m(\X_m), f_{m'}(\X_{m'}))$ from the corresponding kernel $k_{f_m,f_{m'}}(\cdot, \cdot)$ and its hyperparameter $\btheta_{f_m,f_{m'}}$. Given the cross-covariance matrices, the GP prior is placed on $\F$, given by $\F \sim \N(\mathbf{0}, \K)$ where the large covariance matrix $\K$ is comprised of block covariance ($\K_{\Fm,\Fm}$, on the main diagonal blocks) and cross-covariance ($\K_{\Fm,\Fmdash}$ with $m\neq m'$, off-diagonal blocks) matrices. That is,
$$
\F = \begin{pmatrix}
\F_1\\ \vdots \\ \F_M 
\end{pmatrix} \sim \mathcal{N} \left(\mathbf{0}, \begin{pmatrix} \K_{\F_1,\F_1} & \cdots &\K_{\F_1,\F_M} \\ \vdots & \ddots & \vdots \\ \K_{\F_M,\F_1} & \cdots & \K_{\F_M,\F_M} 
\end{pmatrix}\right). 
$$ The MGP is often called exact (or full) when the covariance matrix $\K$ is used without any approximations.

The exact MGP log-likelihood upon integrating over $f_m(\cdot)$ is given as $p(\y | \X_1,\dots,\X_M; \mathbf{\Theta}) = \N(\zero, \K + \bSigma),$ where $\mathbf{\Theta}$ is the collection of all hyperparameters and $\bSigma=\bdiag (\sigma_m^2 \I_{N_m})_{m=1}^M$ with the notation $\bdiag(\mathbf{G}_m)_{m=1}^{M}$ indicating a block diagonal matrix with $\G_1,...,\G_M$ on the main diagonal. Then the process to obtain the estimated hyperparameters and predictive distribution is similar to those of standard GP regression. Note that if the number of observations from each source is the same, e.g., $N_0 = N_m$ for $m=1,\dots,M$, the covariance matrix $\K$ dimension becomes $MN_0 \times MN_0$. The complexity of the exact MGP in this case is $\mathcal{O}(M^3N_0^3)$ which limits its application in practice due to the computation burden. Furthermore, the group membership information is assumed to be fully given for every observation in the MGP construction. This shows that a traditional MGP is infeasible to handle the motivating problem introduced in Sec. \ref{sec::introduction}. 

\section{Proposed Model} 
\label{sec::weak_supervision} 

In this section, we introduce our proposed model for weakly-supervised multi-output regression.

\subsection{Notations} We abuse some notations used in the previous sections to prevent over-complicated notations. We use superscripts $^l$ and $^u$ to indicate quantities associated with ``labeled" and ``unlabeled" observations, respectively.  Here we should note that we define the labeled observation as an observation where a probability for group membership is given.

Consider that we have $N=N^l + N^u$ observations composed of $N^l$ labeled and $N^u$ unlabeled observations. Let $\{ (\x^{l}_n, y^{l}_n) \}_{n=1,...,N^l}$ and $\{ (\x^{u}_n, y^{u}_n) \}_{n=1,...,N^u}$ denote the labeled and unlabeled observations, respectively. For the $n$-th labeled observation we introduce a vector of unknown binary indicators $\z^l_n = [z^l_{nm}]^\top_{m=1,...,M}$, where $\sum_{m=1}^{M} z^l_{nm} = 1$ and $z^l_{nm}\in \{0,1\}$. Let $z^l_{nm} = 1$ indicate that the latent function $\fm (\cdot)$ generated observation $n$. We introduce $\z^u_n = [z^u_{nm}]^\top_{m=1,...,M}$ similarly defined for the unlabeled observations. Along with the indicators, we define a vector of probabilities $\bpi^l_n = [\pi^l_{nm}]^\top_{m=1,\dots,M}$ that represent the probability that the $n$-th observation is generated from source $m$; thus $\sum_{m=1}^{M}\pi^l_{nm} = 1$ for all $n \in \{1,...,N^l\}$. We introduce  similar notation $\bpi^u_n = [\pi^u_{nm}]^\top_{m=1,\dots,M}$ for the unlabeled observations.

For notational clarity in further discussion, we collectively define additional notations as follows: the inputs $\Xu = [\x^{u}_n]^\top_{n=1,...,N^u}$, $\Xl = [\x^{l}_n]^\top_{n=1,...,N^l}$, $\X = [(\Xu)^\top, (\Xl)^\top]^\top$, the observed responses $\y^u = [y^u_n]^\top_{n=1,...,N^u}$, $\y^l = [y^l_n]^\top_{n=1,...,N^l}$, $\Y = [(\y^u)^\top, (\y^l)^\top]^\top$, the group indicators $\Z^u = [\z^{u}_n]_{n=1,...,N^u}^\top$, $\Z^l = [\z^{l}_n]_{n=1,...,N^l}^\top$, $\Z = [(\Z^u)^\top, (\Z^l)^\top]^\top$, the group membership probabilities $\bPi^u = [\bpi^{u}_n]_{n=1,...,N^u}^\top$, $\bPi^l = [\bpi^{l}_n]_{n=1,...,N^l}^\top$, $\bPi = [(\bPi^u)^\top, (\bPi^l)^\top]^\top$, the noise parameters $\boldsymbol{\sigma} = [\sigma_m]_{m=1,...,M}^\top$, the hyperparameters in kernels of an associated GP $\btheta_{(\cdot)}$, and the unknown function outputs $\F$.

\subsection{Framework for Weakly-supervised Multi-output Gaussian Process} 

Now we discuss our model, referred to as weakly-supervised MGP (WSMGP). The general formulation is given as:
\begin{align}
p(\Y | \F, \Z; \boldsymbol{\sigma}) &= \prod_{n,m=1}^{N,M}\N([\Y]_n|[\Fm]_{n}, \sigma_m^2)^{[\Z]_{nm}}, \label{eq:fw1}\\
p(\F | \X; \boldsymbol{\theta}_\F) &= \N(\F |\zero, \K),\label{eq:fw2}\\
p(\Z|\bPi) &= \prod_{n,m=1}^{N,M}[\bPi]_{nm}^{[\Z]_{nm}} = \prod_{n,m=1}^{N^u,M}[\bPi^u]_{nm}^{[\Z]_{nm}}\prod_{n,m=1}^{N^l,M}[\bPi^l]_{nm}^{[\Z]_{nm}}= p(\Z^u|\bPi^u)p(\Z^l|\bPi^l).\label{eq:fw3}
\end{align}

Eq. \eqref{eq:fw1} defines the conditional data likelihood given membership indicators $[\Z]_{nm}$ and Eq. \eqref{eq:fw2} is the Gaussian prior defined over the latent functions that characterizes correlation both \textit{within and between} different outputs through the covariance $\K$.  It is crucial to note that all outputs $1,...,M$ place the prior in terms of all given observations $\X = \X_1 = \cdots = \X_M$, no matter what the given label is. That is, $\X$ concatenates all $N$ input observations and is subject to all outputs $1,...,M$. The membership indicator $[\Z]_{nm}$ then determines whether the group $m$ realizes $[\y]_n$, e.g., the realization $[\y]_n$ for the group $m$ is nullified if $[\Z]_{nm} = 0$. Eq. \eqref{eq:fw3} presents a multinomial distribution parameterized by $\bPi$ for the membership indicator $\Z$. Through the model construction, we can set $\bPi^l$ to the probabilities that represent given knowledge on the label. For example, say we are given that the $n_1$-th observation's label is surely $m_1$, we set $[\bPi^l]_{n_1m_1} = 1$ while $[\bPi^l]_{n_1m} = 0$ for $m\in\{1,...,M\}\backslash \{ m_1 \}$\footnote{In practice, we set $[\bPi^l]_{n_1m_1} = 0.9999$ and $[\bPi^l]_{n_1m} = 0.0001$ for $m\in\{1,...,M\}\backslash \{ m_1 \}$ for numerical stability.}.

For the unlabeled observations $n=1,...,N^u$, we place a Dirichlet prior on $\bpi^u_n$: 
\begin{align}
    p(\bPi^u;\balpha) = \prod_{n=1}^{N^u} \text{Dir}(\bpi^u_n;\balpha) = \prod_{n=1}^{N^u}\left(\frac{1}{\text{B}(\balpha)}\prod_{m=1}^{M}[\bpi^u_n]^{\alpha_0-1}_{m}\right), \label{eq:fw4}
\end{align}
where $\text{B}(\balpha) = \frac{\Gamma^M(\alpha_0)}{\Gamma(M\alpha_0)}$ is a multivariate Beta function with parameter $\balpha = (\alpha_0)_{m=1,...,M}$. In general, the elements of $\balpha$ need not be identical; however, for simplicity, we use identical $\alpha_0$ for all the elements, i.e., a symmetric Dirichlet. Here the Dirichlet prior plays a role in tuning labeling acuteness \citep{bishop2006pattern}.

\subsubsection{Computational Efficiency via Convolved MGPs} As discussed in Sec. \ref{sec::MGP}, the direct application of the exact MGP (full covariance $\K$ in Eq. \eqref{eq:fw2}) induces a significant computational burden. For this reason, we exploit the well-known sparse convolved MGP (SCMGP) \citep{alvarez2011computationally} and extend it to the weakly-supervised setting above. 

Convolved MGPs, a generalization of the linear model of coregionalization \citep{higdon2002space}, construct the latent functions $f_m(\x)$ as convolution of a smoothing kernel $k_m(\x)$ with a latent GP $u(\w)$; $f_m(\x) =\int_{-\infty}^{\infty}k_m(\x-\w)u(\w)d\w$. The only restriction for a valid construction is that the smoothing kernel is absolutely integrable, i.e., $\int|k(w)|dw < \infty$. The key rationale is that if we share the common latent GP $u$ across $M$ sources, then all outputs $f_m$, $m\in\{1,..,M\}$ can be expressed as jointly distributed GP -- an MGP. This construction first proposed by \cite{ver1998constructing} readily generalizes to multiple shared latent GPs \citep{alvarez2011computationally}.

\begin{figure}[h!]
\caption{A plate notation of WSMGP. Larger circles indicate random variables. Smaller squares indicate hyperparameters. Shaded shapes present known values. The half-shaded shape ($\bPi$) indicates that the values of some elements are known.}
\vspace{0.5pc}
	\centering
	\includegraphics[width=0.6\textwidth]{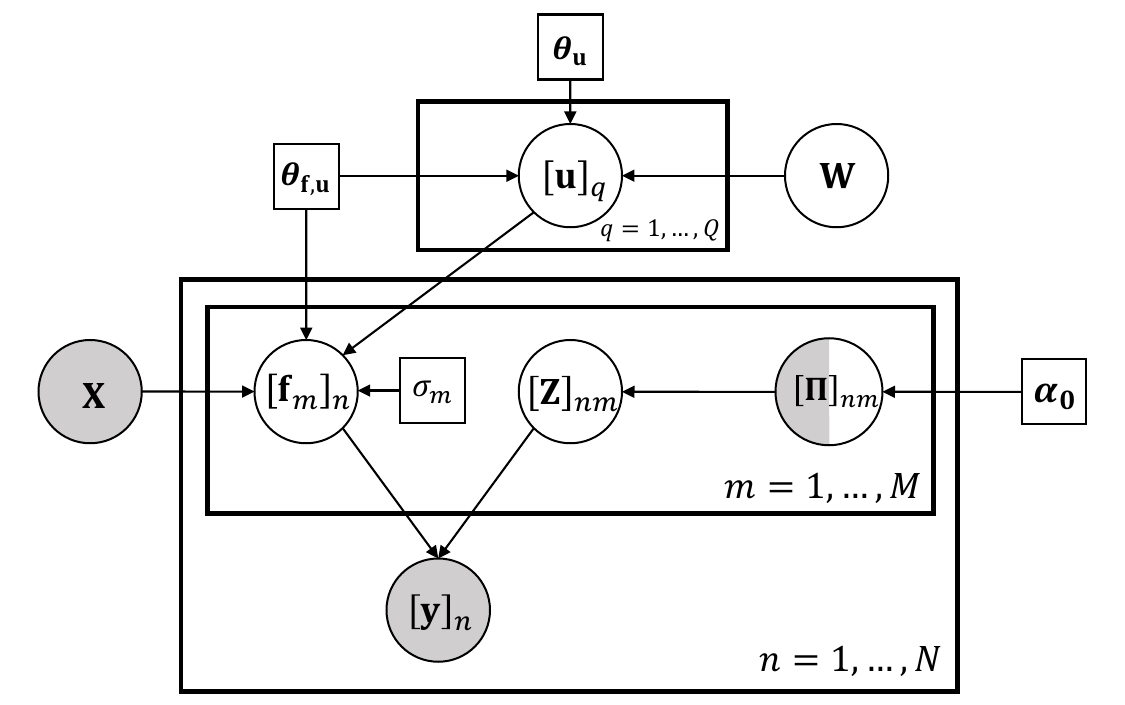}
	\label{fig:plate}
\end{figure}

The advantage of convolved MGPs is that they are amenable to sparse approximations of the large covariance matrix $\K$. Such approximations are based on the fact that all $\F_1,...\F_M$ are independent given $u$. The key assumption is that independence holds if only a set of pseudo-input (also referred to as inducing variables) $\W=\{\w_q\}_{q=1}^{Q}$ are observed from $u$. As a result, the SCMGP approximates $\K$ by substituting Eq. \eqref{eq:fw2} to 
\begin{equation}\label{eq:scmgp}
    p(\F|\W,\X; \btheta_{\F,\bu}, \btheta_{\bu}) = \int p(\F|\mathbf{u},\W,\X; \btheta_{\F,\bu})p(\mathbf{u}|\W; \btheta_{\bu})d\mathbf{u}=\N(\F|\zero, \B + \KFu\K^{-1}_{\mathbf{u},\mathbf{u}}\KuF),
\end{equation}
where $\W=\{\w_q\}_{q=1}^{Q}$ is the set of vectors of inducing points; $\Kuu$ is the covariance matrix of $\bu= [u(\w_1), ...,u(\w_Q)]^\top$ with the hyperparameter $\btheta_\bu$; $\K_{\F,\mathbf{u}} = \K^\top_{\mathbf{u},\F}$ is a cross-covariance matrix with the hyperparameter $\btheta_{\F,\bu}$ relating the column vector $\F=[\F_1^\top, \ldots, \F^\top_M]^\top$ and $\mathbf{u}$ where the column vector $\F_m = f_m(\X)$ collects $f_m$ values at the input $\X$ (recall that we overlapped the inputs $\X_1=\dots=\X_M=\X$). Additionally, $\B = \bdiag(\B_m)_{m=1}^{M}$ with $\B_m = \K_{\Fm,\Fm} - \K_{\Fm,\mathbf{u}}\K^{-1}_{\mathbf{u},\mathbf{u}}\K_{\mathbf{u},\Fm}$, where $\K_{\Fm,\Fm}$ is a covariance matrix of $\Fm$;  $\K_{\Fm,\mathbf{u}} =\K^\top_{\mathbf{u},\Fm}$ is a cross-covariance matrix between $\Fm$ and $\mathbf{u}$. Therefore, $\K$ is approximated to $\B + \KFu\K^{-1}_{\mathbf{u},\mathbf{u}}\KuF$. Fig. \ref{fig:plate} illustrates the relationship of variables in WSMGP with the sparse approximation (i.e., \eqref{eq:fw1}, \eqref{eq:fw3}-\eqref{eq:scmgp}) using a plate notation.

The key idea of Eq. $\eqref{eq:scmgp}$ is that $\K$ is replaced by $\B + \KFu\K^{-1}_{\mathbf{u},\mathbf{u}}\KuF$ which is more computationally efficient when we choose $ Q \ll N $; using Woodbury matrix identity, the computational complexity of the inversion is $\mathcal{O}(N^3 M + NMQ^2)$. Here we should note that we chose convolved GPs as they are well-established low-rank MGP models. However, we should note that our approach allows any sparse approximation to be used within; such as covariance tapering \citep{gramacy2016speeding, kaufman2008covariance}. One can also exploit recent advances in the fast computation of exact GP relying on GPU acceleration and optimized Cholesky decomposition \citep{wang2019exact} or mini-batch approaches \citep{chen2020sgd}. Hereon, we omit the hyperparameters $\{\btheta_{\F,\mathbf{u}}, \btheta_\mathbf{u}, \boldsymbol{\sigma}, \balpha\}$ from the probabilistic models for notational simplicity.

\section{Inference} \label{sec::inference}

Given Eqs. \eqref{eq:fw1}-\eqref{eq:scmgp}, the marginal log-likelihood is written as
\begin{align}
    &\log p(\y|\X,\W) = \log \int_{\F, \mathbf{u}, \Z, \bPi^u} \bigg(  p(\Y | \F, \Z ) p(\F | \mathbf{u}, \X, \W ) p(\mathbf{u}| \W)  p(\Z|\bPi)p(\bPi^u) \bigg),\label{eq:ll}
\end{align}
where $\int_{\mathbf{a}}(\cdot)$ represents integration over the ranges of variables in $\mathbf{a}$. As a result, obtaining a posterior predictive distribution by marginalizing out latent variables in the likelihood is intractable due to the intractability of the marginal likelihood \eqref{eq:ll}. To this end we resort to VI to estimate this posterior. VI approximates the posterior distribution by maximizing an evidence lower bound (ELBO) on the marginal likelihood which is equivalent to minimizing the Kullback-Leibler (KL) divergence $\KL(q(\F,\mathbf{u}, \Z, \bPi) \Vert p(\F,\mathbf{u}, \Z, \bPi |  \X, \y))$ between the candidate variational distributions $q$ and the true posterior \citep{blei2017variational}. Indeed, VI has seen immense success in GPs in the past few years \citep[e.g.,][]{zhao2016variational, panos2018fully}. This success can be attributed to its improved efficiency and generalization capability over other GP inference techniques including sampling based approaches \citep{lalchand2019approximate}, its elegant theoretical properties \citep{yue2019r,burt2019rates}, and VI's ability to be easily fitted within any optimization framework  \citep{hoffman2013stochastic}.    

\subsection{Variational Approximation with Improved Lower Bound}

To proceed we utilize the mean-field assumption in VI,   $q(\F,\mathbf{u}, \Z, \bPi) = p(\F|\mathbf{u})q(\mathbf{u})q(\Z)q(\bPi)$, where $q$ is treated as a family of variational distributions. However one can observe that   the latent variables $\F$ and $\mathbf{u}$ can be analytically marginalized out from Eq. \eqref{eq:ll}. Thus, we only introduce variational distributions over $q(\Z)$ and $q(\bPi^u)$ that belong to the same distributional families as those of the original distributions $p(\Z)$ and $p(\bPi^u)$. We distinguish the variational parameters from the original by using the hat notation (e.g., $\hat \bPi$). 

Our bound can then be derived by observing that $$\log \int_{\Z,\bPi^u} p(\y | \F, \Z) p(\Z| \bPi) p(\bPi^u) \geq \int_{\Z,\bPi^u} q(\Z)q(\bPi^u) \log \frac{ p(\y | \F, \Z) p(\Z | \bPi) p(\bPi^u)}{q(\Z)q(\bPi^u)}$$ as a direct consequence of Jensen's inequality. Exponentiating each side of the above inequality and plugging it into Eq. \eqref{eq:ll}, we find
\begin{equation*}
\begin{split}
    \mathcal{L}_{\text{cVB}} & = \log \int_{\F, \mathbf{u}} \exp{\left(\int_{\Z,\bPi^u} q(\Z)q(\bPi^u) \log \frac{ p(\y | \F, \Z) p(\Z | \bPi) p(\bPi^u)}{q(\Z)q(\bPi^u)}\right)} p(\F | \mathbf{u}, \X, \W) p(\mathbf{u}| \W) \\ & \leq \log p(\y|\X,\W).  
\end{split}
\end{equation*} We refer to this new bound to as the \textit{improved} variational bound $\mathcal{L}_{\text{cVB}}$ due to the fact that it is a tighter bound than the regular variational bound which introduces an additional variational distribution $q(\mathbf{u})$. The integral above can be easily calculated as (refer to Appendix \ref{appdx:der_vb} for a detailed derivation) 
\begin{align}
        \mathcal{L}_{\text{cVB}} = \log \N(\y|\zero, \B + \KFu\K^{-1}_{\mathbf{u},\mathbf{u}}\KuF + \D)+\mathcal{V},\label{eq:Lcvb}
\end{align} 
where $\D = \bdiag(\D_m)_{m=1}^{M}$, $\D_m$ is a diagonal matrix with elements  $\sigma_m^2/[\hat{\bPi}]_{1m},...,\sigma_m^2/[\hat{\bPi}]_{Nm}$ and
\begin{align}
\mathcal{V} = \frac{1}{2} \sum_{n=1,m=1}^{N,M}\log\frac{(2\pi\sigma_m^2)^{(1-[\hat{\bPi}]_{nm})}}{[\hat{\bPi}]_{nm}}-\KL(q(\Z^l)\Vert p(\Z^l | \bPi^l)) -\KL(q(\Z^u)q(\bPi^u) \Vert p(\Z^u | \bPi^u)p(\bPi^u)).
\end{align} Now, the variational parameters $\{ \hat{\bPi}$, $\hat{\boldsymbol{\alpha}} \}$ and hyperparameters $\{ \btheta_{\F,\mathbf{u}}, \btheta_\mathbf{u}, \boldsymbol{\sigma}\}$ can be estimated by maximizing $\mathcal{L}_{\text{cVB}}$. 

\begin{remark}
Note that $\D$ is a diagonal matrix and $\B$ is block-diagonal. Therefore, the computational complexity is $\mathcal{O}(N^3 M + NMQ^2)$. This is a significant reduction in computation compared to using the exact covariance in Eq. \eqref{eq:fw2}. 
\end{remark}

\subsection{Practical Benefits of VI-based Improved Lower Bound} \label{sec:analysis}

In this section, we discuss the practical benefits of our VI framework.   

\subsubsection{Flexibility to Alternative MGPs}\label{sec::ease}

In the lower bound $\Lcvb$, any alternative approximation of $\mathbf{K}$ can be easily utilized in our model. For instance, one may choose to use an exact GP in Eq. \eqref{eq:fw2} instead of the sparse convolved MGP. In this case, the $\Lcvb$ will simply be $\mathcal{L}_{\text{cVB}} = \log \N(\y|\zero, \K + \D)+\mathcal{V}$.

\subsubsection{Interpretable Covariance Matrix} \label{sec::interprete}
Note that $\hat{\bPi}$ is the variational parameter that needs to be optimized along with the parameters of GPs $\{\btheta_{\F,\bu}, \btheta_{\bu}, \bsigma\}$. Obviously, both variational parameter $\hat{\bPi}$ and GP hyperparameters $\{\btheta_{\F,\bu}, \btheta_{\bu}, \bsigma\}$ affect the objective function simultaneously, which potentially hinders efficient optimization. Thus, understanding how the parameters interact is very helpful in designing an efficient optimization algorithm for this problem. 

The matrix $\D$ in the improved lower bound, which is a distinguishing counterpart to the noise matrix $\bSigma$ of a standard SCMGP, controls the influence of an observation to the latent functions using the noise scaled through the variational membership probability. Recall that the diagonals of $\D$ are comprised of $\D_1, ..., \D_M$, corresponding to the GPs $\F_1, ..., \F_M$. Now, suppose that the $n$-th observation is highly likely to be assigned to $\F_{m}$, so $[\hat{\bPi}]_{nm} \rightarrow 1$. Then, we have $\sigma_{m}^2/[\hat{\bPi}]_{nm} \rightarrow \sigma_{m}^2$. This also implies that $[\hat{\bPi}]_{nm'} \rightarrow 0$ for $m' \in \{1,...,M\} \setminus \{m\}$, which yields $\sigma_{m'}^2/[\hat{\bPi}]_{nm'} \rightarrow \infty $. As this noise term goes to infinity the first term value will get closer to zero for any arbitrary input. Thus, it implies that, if $\sigma_{m'}^2/[\hat{\bPi}]_{nm'} \rightarrow \infty $, the likelihood value converges to zero in whatever residual between the observation $n$ and the GP $\F_{m'}$ is. That is, if an observation is perfectly assigned to a certain output, the $\Lcvb$ will be independent of all other outputs.

\subsubsection{Extension to a More Scalable Method for Extremely Large Datasets}\label{sec::scalable}

Despite the reduced complexity of WSMGP, an important extension would be to extend the proposed framework to one that is amenable to stochastic optimization where a mini-batch sampled from the data can dictate the iterative optimization scheme. This will allow scaling to extremely large datasets when needed. However, in GPs, observations are intrinsically correlated and thus $\Lcvb$ cannot be optimized via mini-batches \citep{Hensman2013Gaussian}. To this end, we additionally introduce the variational distribution $q(\mathbf{u})=\N(\mathbf{u}|\bmu_\mathbf{u}, \bS)$ to derive an alternative bound, denoted by $\Lsvb$.
\begin{align} \label{svb}
    &\mathcal{L}_{\textrm{sVB}} = \sum_{n=1, m=1}^{N,M} \mathbb{E}_{q([\Fm]_n)} \left[\log  \mathcal{N}\left([\mathbf{y}]_n \Bigg\vert [\mathbf{f}_m]_n, \frac{\sigma_m^2}{[\hat{\mathbf{\Pi}}]_{nm}}\right) \right] -\KL (q(\mathbf{u})\Vert p(\mathbf{u}))+ \mathcal{V} \, ,
\end{align}
where the approximate marginal posterior for $\Fm$ is given by $q(\Fm) = \int p(\Fm|\mathbf{u})q(\mathbf{u})d\mathbf{u}$ and $q(\Fm) = \mathcal{N}(\Fm | \boldsymbol{\mu}_{q(\Fm)}, \boldsymbol{\Sigma}_{q(\Fm)})$ such that
\begin{align*}
\boldsymbol{\mu}_{q(\Fm)} = \K_{\Fm,\mathbf{u}}\Kuu\boldsymbol{\mu}, \qquad \boldsymbol{\Sigma}_{q(\Fm)} = \K_{\Fm,\Fm} + \K_{\Fm,\mathbf{u}} \Kuuinv(\mathbf{S} - \Kuu)\Kuuinv\K_{\mathbf{u},\Fm}.
\end{align*}

As shown in Eq. \eqref{svb}, the first term of $\Lsvb$ decorrelates the observations, thereby the bound is amenable to stochastic optimization. We provide the detailed derivations of the new bound along with its gradients in Appendix \ref{appdx:der_vb} and \ref{appdx:der_grad}. Although we do not directly pursue the case in this article, we provide $\mathcal{L}_{\textrm{sVB}}$ in case our model is implemented with extremely large data sizes. Such extension and its applications to an extremely large dataset would be a promising future direction to study.

\subsubsection{Controlling Labeling Acuteness by Dirichlet Prior}\label{sec::dirichlet} Recall the second term in $\Lcvb$:
\begin{align}\label{eq:V}
\mathcal{V} = \frac{1}{2} \sum_{n=1,m=1}^{N,M}\log\frac{(2\pi\sigma_m^2)^{(1-[\hat{\bPi}]_{nm})}}{[\hat{\bPi}]_{nm}}-\KL(q(\Z^l)\Vert p(\Z^l | \bPi^l)) -\KL(q(\Z^u)q(\bPi^u) \Vert p(\Z^u | \bPi^u)p(\bPi^u)).
\end{align} 
For the labeled data points, we can easily see that the second term in Eq. \eqref{eq:V} encourages $\hat{\bPi}^l$ to be close to $\bPi^l$ which is the prior label belief. For the unlabeled data points, the hyperparameter $\balpha$ of the Dirichlet prior plays an important role in enforcing regularization. To see this, we first note that the optimal value of variable $\hat{\bbalpha}$, denoted by $\hat{\bbalpha}^*$, is $[\hat{\bbalpha}^*]_{mn} = \hat{\alpha}^*_{nm} = \alpha_{0} + [\hat{\bPi}]_{nm}$. Given $\hat{\bbalpha}^*$, the third term in Eq. \eqref{eq:V} is expressed as
\begin{equation*}
\KL(q(\Z^u)q(\bPi^u)\Vert p(\Z^u | \bPi^u)p(\bPi^u)) = \sum_{m,n=1}^{M,N^u} [\hat{\bPi}]_{nm} \log [\hat{\bPi}]_{nm} - \sum_{n=1}^{N^u} \left( \log \frac{\text{B}(\hat{\alpha}^*_{n1}, ...,\hat{\alpha}^*_{nM})}{\text{B}(\boldsymbol{\alpha}_0)} \right).
\end{equation*}

\begin{figure}[h!]
\caption{$\KL_{\Z,\bPi^u}(\alpha_0)$ is convex or concave in $[\hat{\bPi}]_{n1}$ depending on $\alpha_0$.}
\vspace{0.2in}
	\centering
	\includegraphics[width=0.6\textwidth]{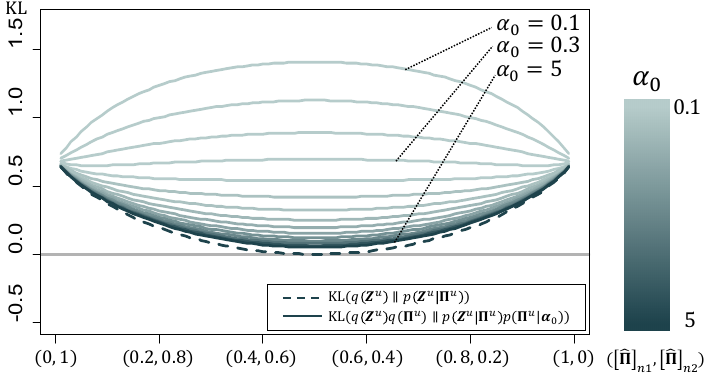}
	\label{fig:KL-alpha}
\end{figure}

Fig. \ref{fig:KL-alpha} demonstrates the above KL-divergence corresponding to the $n$-th observation by $[\hat{\bPi}]_{n1}$ in the case that we consider two GPs: $\sum_{m=1,2} [\hat{\bPi}]_{nm} \log [\hat{\bPi}]_{nm} - \log \frac{\text{B}(\hat{\alpha}^*_{n1},\hat{\alpha}^*_{n2})}{\text{B}(\boldsymbol{\alpha}_0)} = \KL_{\Z,\bPi^u}(\alpha_0).$ Note that small $\KL_{\Z,\bPi^u}(\alpha_0)$ is preferred in maximizing $\Lcvb$. According to the figure, we realize that $\balpha$ controls the \textit{acuteness or discretion} on assignment of observations to sources. To be more detailed, observe that if we set a large $\alpha_0$, (e.g., $\alpha_0 = 5$), then $\KL_{\Z,\bPi^u}(\alpha_0)$ with $([\hat{\bPi}]_{n1}, [\hat{\bPi}]_{n2}) = (0,1)$ or $(1,0)$ is greater than the one with $([\hat{\bPi}]_{n1}, [\hat{\bPi}]_{n2})= (0.5,0.5)$. On the other hand, if we have a small $\alpha_0$ (e.g., $\alpha_0=0.1$), $\KL_{\Z,\bPi^u}(\alpha_0)$ with $([\hat{\bPi}]_{n1}, [\hat{\bPi}]_{n2}) = (0.5,0.5)$ is greater than the one with $([\hat{\bPi}]_{n1}, [\hat{\bPi}]_{n2})= (1,0)$ or $(0, 1)$. Also, $\KL_{\Z,\bPi^u}(\alpha_0)$ converges to   $\KL(q(\Z^u)\Vert p(\Z^u | \bPi^u))$ as $\alpha_0$ increases. 

\section{Prediction}
\label{sec::prediction}
In prediction, we are interested in the posterior predictive distribution of the $m$-th output at a new input $\x^*$. Given the optimized variational parameters $\{ \hat{\bPi}^*$, $\hat{\boldsymbol{\alpha}}^* \}$ and hyperparameters $\{\btheta_\F^*, \btheta_{\F,\mathbf{u}}^*, \btheta_\mathbf{u}^*, \boldsymbol{\sigma}^*\}$, the distribution is derived by marginalizing out the likelihood at $\x^*$ with respect to the latent variables using their posterior distributions. In the VI framework, we substitute the intractable exact posterior distributions with the corresponding approximated posterior distributions. 

 We first derive the posterior distribution of $\mathbf{u}$ by applying Bayes' theorem to $p(\y|\mathbf{u}, \X, \W)$ and $p(\mathbf{u}|\W)$, expressed as $$ p(\mathbf{u}|\y, \X, \W) = \N(\Kuu \mathbf{A}^{-1}\KuF (\B+\D)^{-1}\y, \Kuu \mathbf{A}^{-1}\Kuu) $$ where $\bA = \Kuu + \KuF (\B + \D)^{-1}\KFu$. Then the posterior predictive distribution of the $m$-th output is expressed as 
\begin{align*}
&p(\y_m^*|\x^*, \y, \X, \W) \\
 &\quad= \int_{\F, \mathbf{u}, \Z, \bPi^u} p(\y_m^* | \Fm, \Z, \x^*, \y, \X, \W) p(\Fm|\mathbf{u},\y, \X)p(\mathbf{u}|\y,\X,\W)p(\Z|\y,\X)p(\bPi^u|\y,\X)\\
 &\quad\approx \int_{\F, \mathbf{u}, \Z, \bPi^u} p(\y_m^* | \Fm,\Z, \x^*, \y, \X, \W) p(\Fm|\mathbf{u},\y, \X)p(\mathbf{u}|\y,\X,\W)q(\Z)q(\bPi^u)\\
 &\quad\approx\N(\K_{\Fm^*,\mathbf{u}}\mathbf{A}^{-1}\K_{\mathbf{u},\Fm} (\B_m + \D_m)^{-1}\y, \B^*_m + \K_{\Fm^*,\mathbf{u}}\mathbf{A}^{-1}\K_{\mathbf{u}, \Fm^*} + \sigma^2_m \I),
\end{align*} where $\B^*_m = \K_{\Fm^*,\Fm^*} - \K_{\Fm^*,\mathbf{u}}\K^{-1}_{\mathbf{u},\mathbf{u}}\K_{\mathbf{u},\Fm^*}$, with $\K_{\Fm^*,\Fm^*}$ and $\K_{\Fm^*,\mathbf{u}} = \K_{\mathbf{u},\Fm^*}^\top$ which are the matrices defined similarly to the (cross-) covariance matrices in SCMGP, but evaluated at $\x^*$.

\section{Related Work}
\label{sec::related_work}

Multi-output regression has been an active research area for its applicability to real-world problems that involve multiple outputs. Examples can be found in a wide range of applications from engineering to medical science, such as modeling multiple quality measurements of manufactured parts or multivariate vital-sign analysis, to name a few. In the methodological context, multi-output regression models can be categorized into problem transformation and algorithm adaptation methods \citep{borchani2015survey}. Problem transformation approaches model each output independently and finally concatenate the predictors from all the outputs. While computationally efficient, the main drawback of such methods is that they ignore the potential correlation between outputs. On the contrary, algorithm adaptation approaches seek to capture dependencies across outputs. As a result, outputs can borrow knowledge from each other, often resulting in improved predictions. In this regard, our model falls into the category of algorithm adaptation methods. 

Among various algorithm adaptation methods, statistical models have gained significant popularity as they can provide a principled approach to modeling correlation. These approaches extend a classical regression model with a single response to a model with multivariate responses, treating them as a vector-valued variable. Early work on statistical treatments for multivariate responses can be seen in reduced-rank regression \citep[e.g.,][]{izenman1975reduced}, partial least-squares regression \citep[e.g.,][]{wold1975soft}, and multivariate regularized regression \citep[e.g.,][]{brown1980adaptive}. Along this line, \cite{breiman1997predicting} proposed the ``Curds-and-Whey'' method that directly exploits the potential correlation across response variables through estimating a shrinkage matrix multiplied to the ordinary least squares predictors. More recent studies have also investigated multivariate regression through sparse regularization \citep[e.g.,][]{rothman2010sparse, li2015multivariate}. Here we should note that the above methods for vector-valued responses make assumptions that limit their applicability. Examples of such assumptions are that all outputs have the same input and the predicted curves follow a pre-specified parametric form. To remedy these issues, non-parametric approaches that allow different input across outputs have been explored. Some examples are support vector regression \citep[e.g.,][]{sanchez2004svm, xu2013multi} or regression tree approaches to multi-output models \citep[e.g.,][]{jeong2020regularization}. Such methods often bring advantages in predictive accuracy compared to the traditional statistical models, yet the lack of explainability and uncertainty quantification still hinders their broad use. 

To address the above shortcomings, recent research has paid increased attention to MGPs as an elegant treatment to multi-output regression. Indeed, GPs are naturally extended to an MGP, as all outputs can be simply considered as one observation from a single GP with both within and across output correlation. Besides that, MGP offers the inherent capability to quantify uncertainty and avoid parametric assumptions. Such advantageous features have made MGP a popular approach in a variety of studies that deal with multi-output regression. In particular, the spatial statistics community has intensively studied MGPs, often referred to as co-kriging \citep[e.g.,][]{ver1998constructing, chiles2009geostatistics}. Studies have proposed efficient covariance structures that account for complex spatial features \citep[e.g.,][]{furrer2011aggregation,genton2015cross} and applied such models in a variety of spatial statistics applications  \citep[e.g.,][]{nerini2010cokriging, adhikary2017cokriging, bae2018missing}. MGPs are also investigated in the field of computer experiments where an MGP acts as a surrogate model for expensive physical experiments \citep[e.g.,][]{qian2008gaussian, fricker2013multivariate, deng2017additive}. Another field of applied statistics that successfully employs MGPs is reliability engineering \citep[e.g.,][]{kontar2017nonparametric, kontar2018nonparametric, jahani2018statistical}, where the remaining useful life of multiple correlated system components are integratively analyzed.  Meanwhile, the machine learning community has extensively focused on approximate inference for scaling to large datasets. Amongst the proposed methods are: sparse MGPs \citep[e.g.,][]{alvarez2009sparse}, distributed MGPs \citep[e.g.,][]{kontar2020minimizing}, VI for low rank inference of MGPs \citep[e.g.,][]{alvarez2010efficient, zhao2016variational, moreno2018heterogeneous}, GPU acceleration \citep[e.g.,][]{wang2019exact} and structured kernel inversions \citep[e.g.,][]{saatcci2012scalable, wilson2015kernel}.

While various attempts on both theory and applications were made, a common assumption across MGP literature is that group labels are readily available. Unfortunately, this is not true in many practical applications. The few papers that address missingness in GPs focus on classification tasks \citep[e.g.,][]{lawrence2005semi,skolidis2013semisupervised, damianou2015semi}. From a regression perspective, the works are numbered. \citet{Jean2018Semisupervised} proposed an approach based on deep kernel learning \citep{wilson2016deep} for regression with missing data, inferring a posterior distribution that minimizes the variance of unlabeled data as well as labeled data. \citet{cardona2015convolved} modeled a convolved MGP for semi-supervised learning. This literature however defines missing labels as observations of which the responses $y_{mn}$ are missing. In contrast, our study considers weakly-supervised learning for the case of missing labels for group memberships rather than missing responses.

In this sense, our study is most related to \citet{lazaro2012overlapping} who solve data association problem using overlapping mixture of Gaussian processes (OMGP). The goal of data association is to infer the movement trajectories of different objects while recovering the labels identifying which trajectory corresponds to which object. Extending the OMGP, \citet{ross2013nonparametric} proposed a model that enables the inference of the number of latent GPs. Recently, \citet{kaiser2018data} also extended the OMGP by modeling both the latent functions and the data associations using GPs. However, OMGP makes a key assumption of independence across mixture components (i.e., the outputs). This limits its capability in scenarios that call for modeling between-output correlations and borrowing strength across outputs, a key feature we incorporate in this work.

\subsection{A Comparison to Fully-supervised and Unsupervised Models}

Weak-supervision is due to our model's ability to exploit both labeled and unlabeled observations, where the ``label'' stands for group membership.  In this sense, an unsupervised and fully-supervised GP for multi-output regression can be compared to our model. We here discuss OMGP \citep{lazaro2012overlapping} and SCMGP \citep{alvarez2011computationally} as an unsupervised alternative and a fully-supervised alternative, respectively.

OMGP is designed to uncover underlying functions of observations generated from different sources. It mainly focuses on unsupervised learning tasks where no data points have labels. On the other hand, WSMGP is applicable to cases where all data points have group membership labels, no data points have labels, and some data points have labels, while allowing the labels to be probabilistic and to reflect a prior belief on the label. In addition, WSMGP deals with multi-output regression. Our goal is to borrow strength across different outputs/groups to improve their learning and predictive accuracy. The knowledge transfer is done through the latent process $u$ shared by the outputs $f_1,...,f_M$. In contrast, OMGP assumes no dependencies between the curves. Indeed, OMGP can easily fail if curves have commonalities due to its construction (as shown in our experiments). 

SCMGP can be viewed as a fully-supervised learning approach for multi-output regression. It only uses labeled observations whose group memberships are known. Consequently, SCMGP struggles when only a few observations have labels. On the contrary, WSMGP can excel even with less labeled observations by using knowledge from unlabeled observations. This is enabled through the overlapping mixture of correlated GPs with the membership indicators $\Z$. Besides, SCMGP and WSMGP are distinguished in terms of model inference. SCMGP has a closed form of the marginal log-likelihood \eqref{eq:scmgp} which leads to maximum likelihood estimation for model inference. However, deriving a closed-form of the marginal log-likelihood of WSMGP is not tractable, inducing the need for an alternative approach such as VI.

\section{Experimental Results}
\label{sec:experiment}
We assess the performance of the proposed model using both synthetic and real-world healthcare data. We assume all given labels are accurate. We compare WSMGP to four benchmark models, two OMGP-based and two SCMGP-based benchmarks, as follows.
\begin{itemize}
    \item[(i)] OMGP (\cite{lazaro2012overlapping}), where $[\bPi]_{nm} = 1/M$, for any $n,m$. This approach ignores given labels.
    
    \item[(ii)] OMGP-WS, a weakly-supervised OMGP approach, where $[\bPi]_{nm'} = 1$ and $[\bPi]_{nm} = 0 $ for $m\in\{1,...,M\}\setminus \{m'\}$ if the label $m'$ is given for the $n$-th observation.
    
    \item[(iii)] SCMGP (\cite{alvarez2011computationally}), using only labeled observations. Comparing WSMGP with SCMGP will show WSMGP's ability to leverage unlabeled data.
    
    \item[(iv)] SCMGP-LP, a two-step approach that first infers group labels and then infers curves using SCMGP. The Label Propagation algorithm by \cite{zhu2002learning} is employed for the first step. Comparing WSMGP with SCMGP-LP will highlight the benefit of joint inference in WSMGP for labels and curves.
\end{itemize}

Table \ref{tab:benchmarks} summarizes the proposed model and benchmarks. All experiments are run on an Intel(R) Core(TM) i7-8700 CPU @ 3.40GHz (8 CPUs) and 32GB RAM, implemented in \textbf{\textsf{R}} version 3.5.0.

\begin{table}[h]
	\centering
	\caption{Comparative models and their considerations in the experiments.}\vspace{0.2cm}
	\adjustbox{max width=\textwidth}{\begin{tabular}{c|c|c|c|c}
			\hline \multirow{2}[0]{*}{Model} & \multicolumn{4}{c}{Model considerations} \\
			\cline{2-5}          & Correlation & Labeled  & Unlabeled & Joint inference \\
			\hline
			WSMGP & \checkmark     & \checkmark     &  \checkmark & \checkmark   \\
			OMGP-WS & \xmark     & \checkmark      & \checkmark & \checkmark      \\
		    OMGP  & \xmark     & \xmark     & \checkmark & \checkmark      \\
			SCMGP & \checkmark      & \checkmark      & \xmark  & \xmark    \\
			SCMGP-LP & \checkmark      & \checkmark      &  \checkmark & \xmark \\
			\hline
	\end{tabular}}%
	\label{tab:benchmarks}%
\end{table}%

\subsection{Simulation Studies}
\label{sec:synthetic}

For WSMGP, we use one latent process $u(\w)$ modeled as a GP with squared exponential kernel $$k_{u,u}(\w,\w') = \exp \left[ -\frac{1}{2}(\w-\w')^T\bL(\w-\w') \right],$$ where $\bL$ is a diagonal matrix. We also use the smoothing kernel 
$$k_m(\x-\w) = \frac{S_{m}|\bL_m|^{1/2}}{(2\pi)^{p/2}} \exp \left[ -\frac{1}{2} (\x-\w)^T \bL_m (\x-\w)  \right]$$ with $S_{m} \in \mathbb{R}$ and a positive definite matrix $\bL_m$, which is widely studied in the literature \citep[e.g.,][]{alvarez2011computationally, alvarez2019nonlinear}. As benchmarks, a squared exponential kernel is used for each independent GP. The data is generated from an MGP composed of two GPs corresponding to the sources (e.g., $m=1,2$), with a kernel $k_{f_m,f_m}(\x, \x')$ where $\bL_1 = 120, \bL_2 = 200, S_1 = 4, S_2=5, \bL=100,$ $\sigma_1 = \sigma_2 = 0.25$ and $Q=30$. Note that the matrices $\bL_1$, $\bL_2$, $\bL$ $\in \mathbb{R}^1$ as the input dimension $d=1$.  For each source we generate 120 observations. As in the motivating example, we consider sparse observations from source 1. We introduce ``$\gamma$-sparsity'' to indicate the ratio of the number of observations from source 1 to those from source 2. In addition, we use ``$l$-dense" to indicate that $l$ fraction of the observed data from each source are labeled. Finally, we set $\alpha_0= 0.3 $ for the Dirichlet prior. We found that this setting works reasonably well in general, though it can be  included in the parameters to be optimized. 


\textbf{Imbalanced populations} We first investigate the behavior of WSMGP when we have imbalanced populations. To do this, we set $\gamma=0.2,0.3,0.5$ and $l=0.2,0.3,0.5$ to mimic distinct levels of imbalance and fractions of the observed labels. We evaluate the prediction performance of each model by the root mean squared error (RMSE), where the errors are evaluated at $\x^*$. To clearly separate the two groups/sources, in the simulations, we add a constant mean of $b=2$ to the GP in source 2.

\begin{figure*}[t!]
\caption{An illustrative example for two imbalanced populations where ({\sf imbalance ratio, labeled fraction}) $=$ $(\gamma, l) = (0.3, 0.2)$. Note that the observations from source 1 are sparse. The observed data points and labels are colored in the first panel; The data is fitted by all models where the inferred probability of each observation's label is represented by a color gradient. Best viewed in color.}
\vspace{0.2in}
\begin{center}
		\centerline{\includegraphics[width=\textwidth]{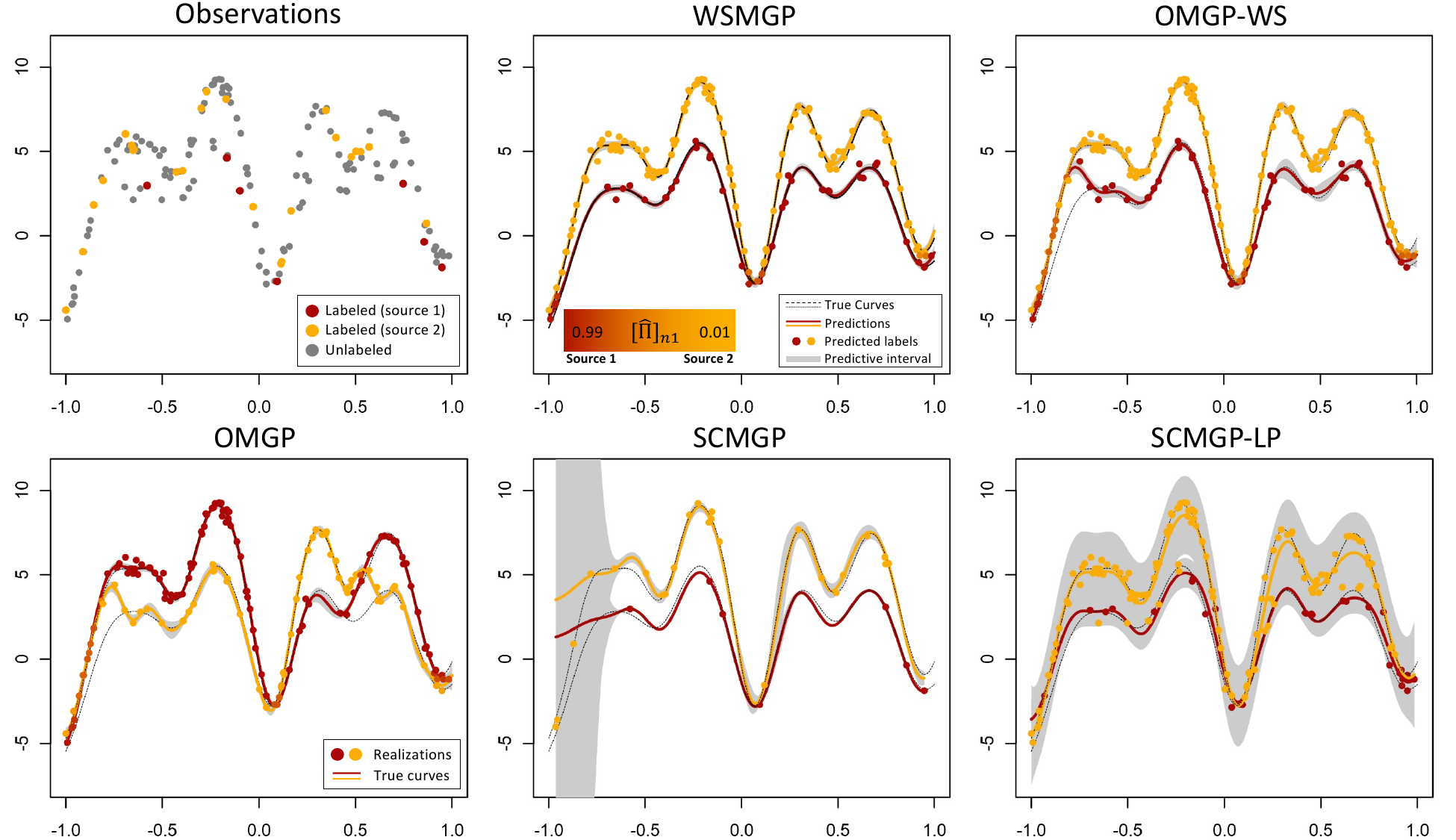}}
		\label{fig:impop}
	\end{center}
\end{figure*}

\begin{figure}[h!]
	\caption{Average RMSEs. Standard deviations are omitted for clear representation.}
	\vspace{0.2in}
	\centering
	\includegraphics[width=0.7\textwidth]{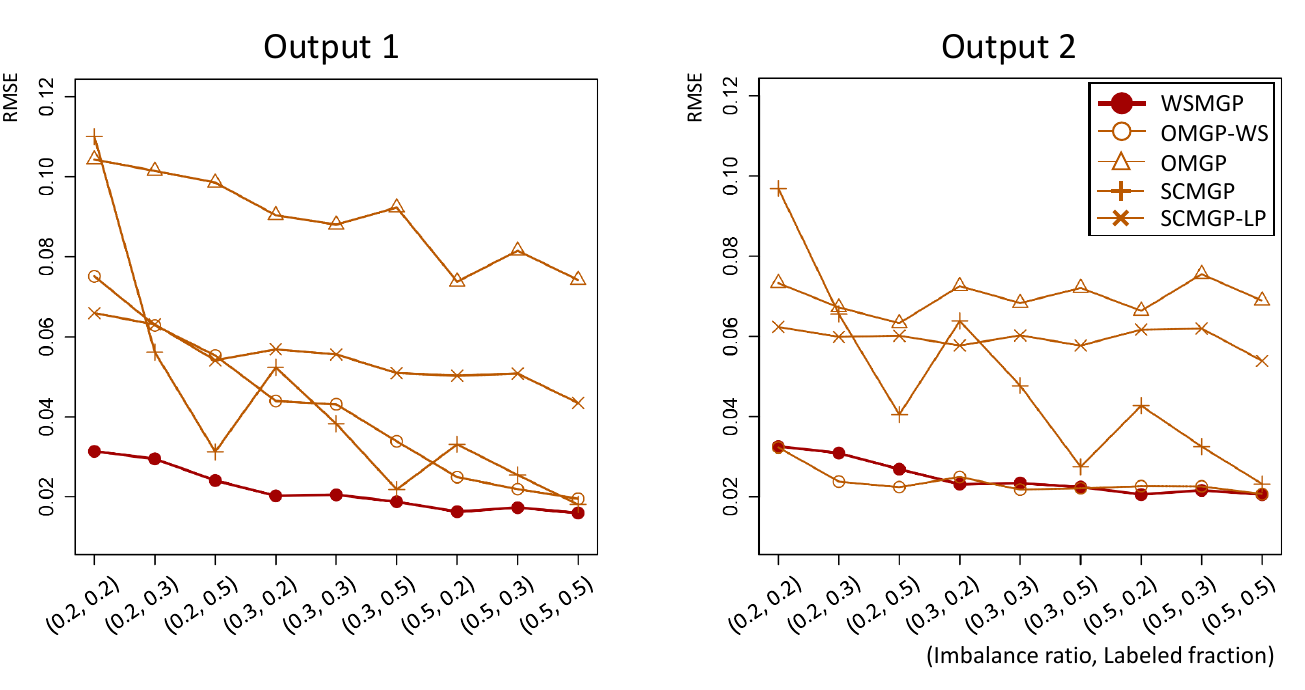}
	\label{fig:mse}
\end{figure}

We make five key observations about the results shown in Fig. \ref{fig:impop} and \ref{fig:mse}. First, the WSMGP model outperforms the benchmarks, especially for source 1 at low $\gamma$ values (highly imbalanced populations). Therefore, WSMGP can use the information of the dense group (source 2) for predicting the responses in the sparse group (source 1). In contrast, the benchmarks model the latent curves independently resulting in poor predictions for the sparse group. Importantly, WSMGP infers the label of each observation with an overall higher accuracy relative to the benchmarks. Both better prediction and more accurate label recovery are direct benefits offered by our model's capacity to incorporate the between-group correlations. Second, Fig. \ref{fig:mse} shows that the prediction accuracy of each model improves as the population becomes more balanced with more labels (larger values of $\gamma$ and $l$). Specifically, we did not observe significant reductions in RMSE for OMGP as the fraction of observed labels $l$ increases. This is not surprising given OMGP ignores the labels.
Third, the WSMGP model outperforms the SCMGP model in both groups. This result illustrates the capability of WSMGP for weakly-supervised learning, which makes use of information from unlabeled data to improve predictions.  Fourth, the WSMGP model has small RMSEs for both sources, while the OMGP-WS model cannot predict well for the sparse group 1 (Fig. \ref{fig:mse}). For example, with enough number of labels or observations, the OMGP-WS can make a reasonable prediction for the dense group (see the prediction for source 2 in the panel ``OMGP-WS'' in Fig. \ref{fig:impop}), while the predictions for the sparse group is poor. The ability of WSMGP to predict the responses well, even for sparse groups, results in a similar level of test accuracies between the two groups. Such results highlight the potential of our model in achieving \textit{group fairness}, defined by a function of discrepancy (e.g., variance) in test accuracies across groups \citep[e.g.,][]{barocas2017fairness, mehrabi2021survey, yue2021gifair}. Indeed, this is a promising result because unlabeled and unbalanced data settings may often arise in the presence of minority groups in practice. We will see real-world examples in the case studies. Improving our model to promote group fairness further is an exciting direction to pursue in future work. Fifth, WSMGP consistently provides more accurate predictions than SCMGP-LP. The result shows the advantage of joint modeling compared to a two-step approach that infers labels and then curves sequentially \citep{yue2021joint}. Furthermore, even with exploiting unlabeled data, SCMGP-LP does not always outperform SCMGP which completely ignores unlabeled data. This is understandable, as SCMGP-LP is vulnerable to inaccurate label estimation in the first step. This is further highlighted in the following experiment.

\begin{figure*}[h!]
\caption{An illustrative example with overlapping and similar curves. We make the green group sparse and set ({\sf imbalance ratio, labeled fraction}) $=$ $(\gamma, l) = (0.3, 0.3)$.}
\vspace{0.2in}
\begin{center}
\centerline{\includegraphics[width=\textwidth]{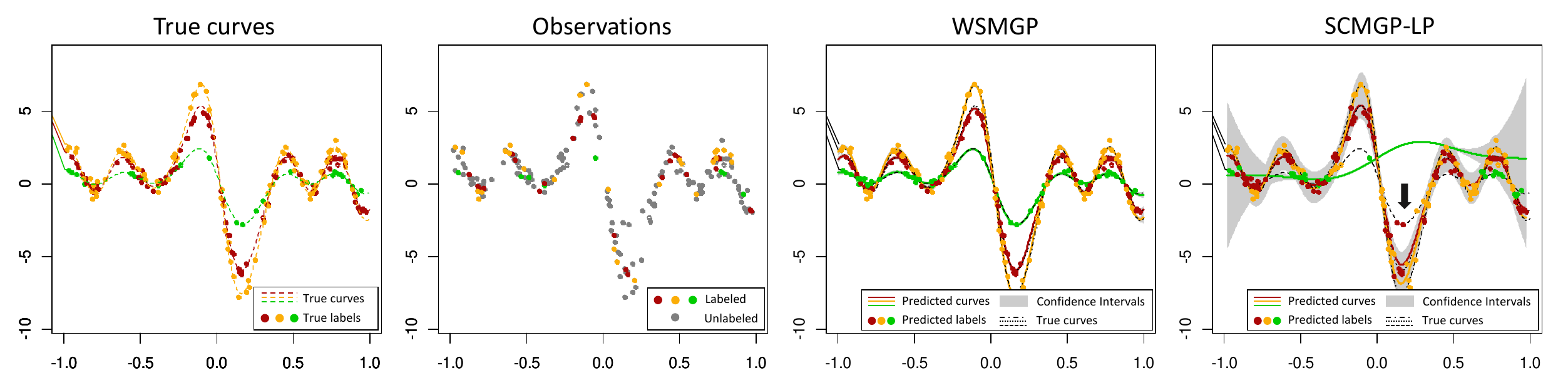}}
\label{fig:sim}
\end{center}
\vspace{-0.2in}
\end{figure*}

\textbf{Revealing similar latent functions} In this experiment, we generate data with three outputs (e.g., $m=1,2,3$) where two of them are very similar; we set $\bL_1 = 120, \bL_2 = 180, \bL_3 = 200, S_1 = 2, S_2=4, S_3=5, \bL=100, \sigma_1 = \sigma_2 = \sigma_3 = 0.25$, $Q=30$, and $b=0$. Similar to the previous experiment we make one of the groups sparse and set sparsity and partially observed labels to $(\gamma, l) = (0.3, 0.3)$. An illustrative example is represented in the second panel of Fig. \ref{fig:sim}. We remark that, based on the observations in the second panel, it is very difficult to distinguish the true curve of each source by a human observer. The differences between similar curves mainly come from the peaks and valleys (e.g., $[-0.2,0]$), and the WSMGP performs well to reveal the underlying similar curves in those intervals. 

In addition, Fig. \ref{fig:sim} highlights a key advantage of WSMGP regarding joint inference for labels and curves.
SCMGP-LP, a straightforward two-step approach instead of joint inference, may suffer immensely from error propagation, specifically when labels are poorly predicted. This issue will be especially important in the presence of correlation and overlap across outputs which make their separation more challenging. Fig. \ref{fig:sim} showcases how the two-step approach can fail. In the first step, the approach makes a misleading label prediction for the observations indicated by the black arrow in the fourth panel, which should have been assigned to the green group. The observation is due to the apparent overlap across curves where nearby labels belong to the red or yellow group. The wrong label predictions eventually lead to poor curve prediction with high variance in the second step. On the contrary, WSMGP, where the label and curve estimation are done simultaneously, can address this challenge.

\begin{figure}[h!]
\caption{Predictions by WSMGP and WSMGP without Dirichlet prior. Plots in the second row illustrate the posterior probability of belonging to the source 1 for each observation ($[\hat{\bPi}]_{n1}$).}
\vspace{0.2in}
\label{fig:Dir}
	\centering
	\includegraphics[width=0.7\textwidth]{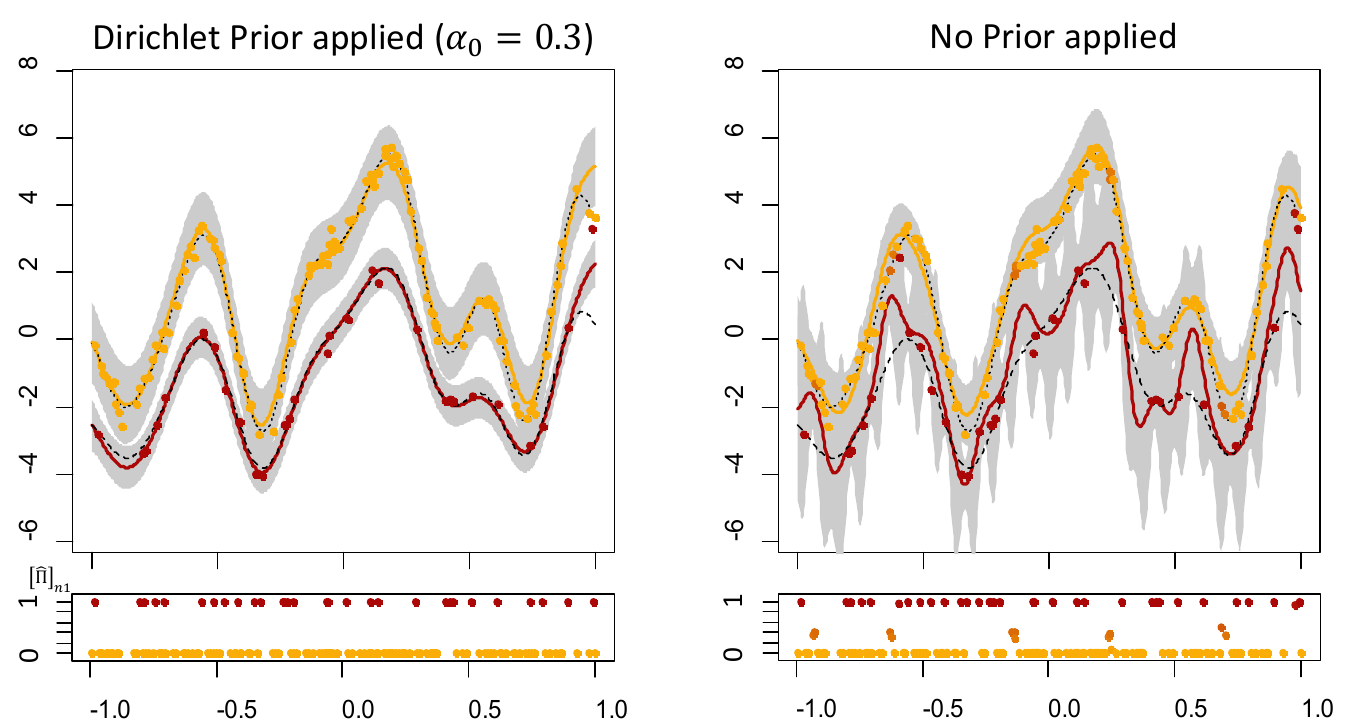}
\end{figure}

\textbf{Dirichlet prior} In this experiment we investigate the effect of varying the hyperparameter $\alpha_0$ of the Dirichlet prior. We compare the WSMGP with $\alpha_0 = 0.3$ and an alternative model, WSMGP-NoDir, obtained by removing the Dirichlet prior from WSMGP. The result is demonstrated in Fig. $\ref{fig:Dir}$. Note that the plots in the second row represent $[\hat{\bPi}]_{n1}$, with values close to 1 or 0 indicating high levels of posterior certainty of the inferred labels.

From the second row of the figure, we can find that the WSMGP-NoDir does not assign proper probabilities to some observations which leads to a poor prediction, while WSMGP perfectly assigns the probabilities for every observation. This is because the inference of $\hat{\bPi}$ depends on the Dirichlet prior. Specifically, as we discussed in Sec. \ref{sec:analysis}, if a small $\alpha_0$ is given (e.g., $\alpha=0.3$), the model is encouraged to find a probability that clearly specifies a group for unlabeled observations, and vice versa. In particular, if $\alpha_0$ is large enough (e.g., $\alpha_0$=100), the WSMGP converges to WSMGP-NoDir. This shows that placing a Dirichlet prior enables WSMGP to achieve more flexibility by searching for a proper $\alpha_0$, while WSMGP-NoDir does not have this option.

\subsection{Case Study I: Insulin Data}
\label{sec::real_data_analysis}

In this section, we discuss the real application of the proposed model that estimates the functional relation between insulin level and the C-peptide level in the blood. Insulin is a hormone that controls blood sugar levels in the body.  C-peptide is a substance excreted when insulin is produced. It is often considered a better biomarker than insulin, as the pancreas excretes C-peptide at a more constant rate and stays in the body longer than insulin. Therefore, it is practically important to know how much C-peptide is produced given a certain insulin level \citep{guildford2020can}. Here we have two challenges for predicting C-peptide levels. First, many people did not disclose whether they take insulin or not, i.e., observations with missing group labels. Second, due to the scarcity in the data for those taking insulin, it is very challenging to learn an accurate predictive model for C-peptide without borrowing knowledge from the dense group not taking insulin. WSMGP can offer an elegant solution that allows a correlated multi-output treatment where groups borrow strength from each other, as well as the ability to borrow strength from observations with no labels.


We obtain data, which we denote as \textit{Insulin} dataset, from the National Health and Nutrition Examination Survey (NHANES) conducted by the National Center for Health Statistics\footnote{\url{https://www.cdc.gov/nchs/nhanes/}} in 1999-2004. In the NHANES database, the insulin and C-peptide measurements are denoted by LBXINSI and LBXCPXI, respectively. We divide the observations into two groups based on the questionnaire about insulin use: a group taking insulin is denoted by `Y' and the other group that does not take insulin is denoted by `N'. The number of total observations is 9418, where the number of people in the group Y and N is 65 and 8478, respectively. The remaining 874 people did not answer the questionnaire and, hence, correspond to the observations with missing labels. To form a training dataset, we randomly choose about 20\% of the group Y, 1\% of the group N and 20\% of the group of missing labels. We note that the training data features highly unbalanced populations across groups where the number of observations from group $Y$ is much less than from group $N$. 
The experiment using such training data will show how WSMGP and the benchmark models work under unbalanced populations. The test data is composed of the remaining labeled observations. Note that we cannot use the unlabeled observations as test data because we cannot compare them with their true label. We repeat the above procedure 10 times each with randomly selected data points.

\begin{figure*}[h!]
\caption{Illustration of the predicted curves with training data, predicted labels, and test data for the Insulin dataset. Note that WSMGP, OMGP-WS, and SCMGP-LP use both labeled (colored) and unlabeled (grey) data. OMGP uses the values of both labeled and unlabeled observations, but does not use label information. SCMGP uses labeled data only. Best viewed in color.}
\vspace{0.2in}
\begin{center}
\centerline{\includegraphics[width=\textwidth]{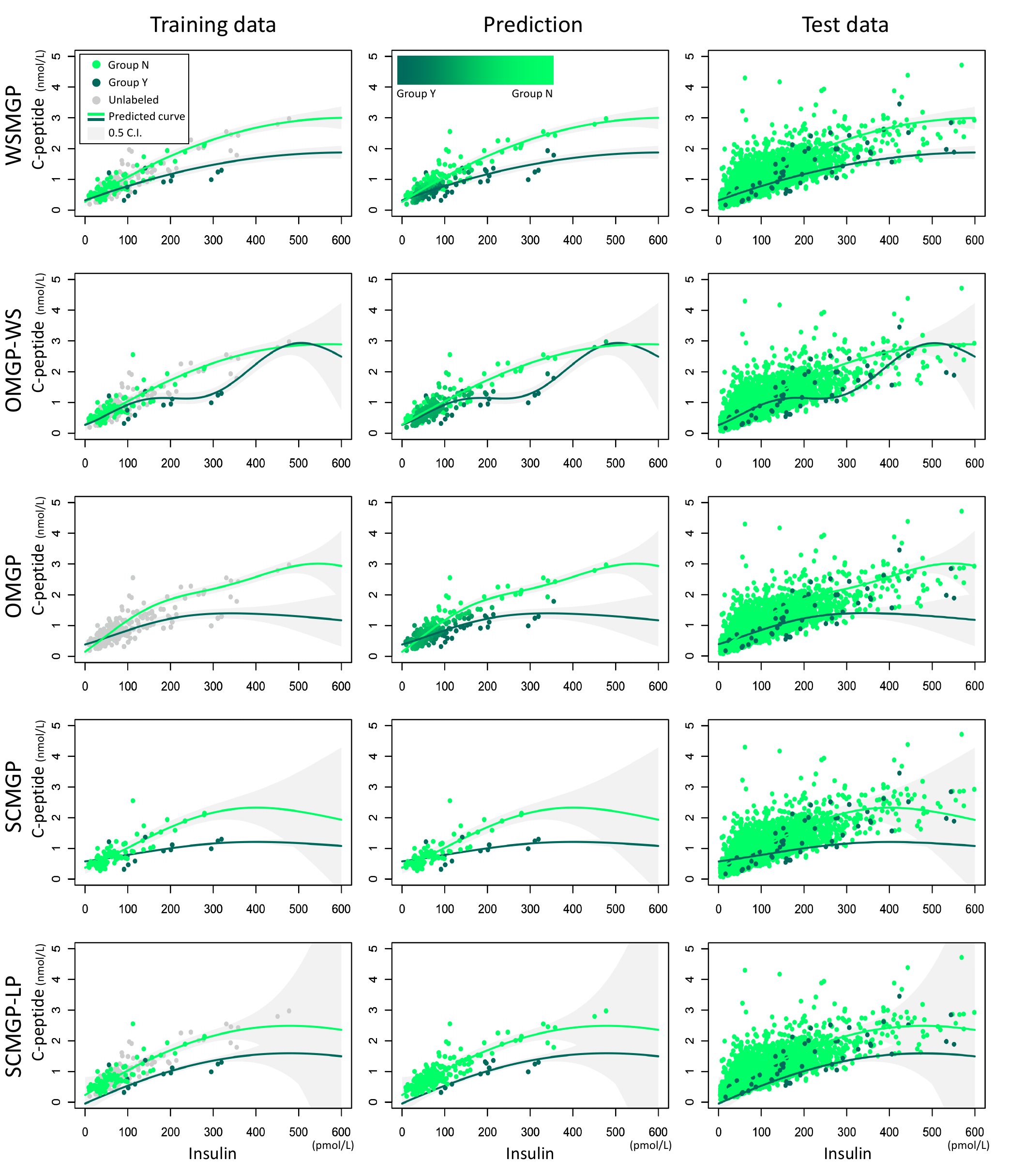}}
\label{fig:insulin-plots}
\end{center}
\vskip -0.3in
\end{figure*}

Experimental results using the Insulin dataset are provided in Table \ref{tab:insulin-mse} and Fig. \ref{fig:insulin-plots}. From the results, we observe that WSMGP attains better prediction accuracy compared to the benchmark models. In particular, WSMGP outperforms SCMGP, which again confirms that exploiting information from the unlabeled observations is critical for accurate predictions under weakly-supervised settings. We see that SCMGP-LP can improve the performance of SCMGP by exploiting unlabeled data. However, WSMGP still provides better results than SCMGP-LP due to the joint inference of curves and labels. Furthermore, WSMGP performs significantly better than OMGP-WS and OMGP in group Y of which observations are sparse. This result highlights the advantages of using WSMGP to deal with an unbalanced population; WSMGP effectively extrapolates the curve for the sparse group by leveraging latent correlations between the groups. On the other hand, OMGP-WS and OMGP that assume independent outputs provide good predictions for the dense group N but misguided predictions for the sparse group Y. The illustration of predicted curves for each model in Fig. \ref{fig:insulin-plots} supports this observation. Note that the benchmark models predict the curve for the dense group N quite well. This is explained by the fact that the given observations of group N are sufficient to characterize the latent curve, as represented in Fig. \ref{fig:insulin-plots}.

\begin{table}[h!]
  \centering
  \caption{Averages and standard deviations of RMSEs from 10 repeated experiments for the test data of Insulin dataset. The best results are boldfaced. Computation times are in seconds. }
  \vspace{0.2in}
    \begin{tabular}{c|ccccc}
    \hline
    \multirow{2}[0]{*}{Group} & \multicolumn{5}{c}{Average RMSE (standard deviation)} \\
\cline{2-6}          & WSMGP & OMGP-WS & OMGP  & SCMGP & SCMGP-LP \\
    \hline
    Y     & \textbf{0.50} (0.04) & 0.67 (0.27) & 0.79 (0.20)     & 0.59 (0.12) & 0.55 (0.24) \\
    N     & \textbf{0.32} (0.06) & \textbf{0.32} (0.03) & 0.35 (0.05) & 0.35 (0.12) & 0.35 (0.10) \\
    \hline
    Elapsed time (total) & 139.05 &	139.75	&143.86&	82.14 &  90.29

 \\
    Elapsed time (per iteration) & 0.2579&	0.2891 &	0.2588 &	0.2106 &  0.2135

\\
    \hline
    \end{tabular}%
  \label{tab:insulin-mse}%
\end{table}%

\subsection{Case Study II: Testosterone Data}

In this section, we apply our model to predict testosterone levels in males. Testosterone is a sex hormone that affects the male reproductive system and physical sturdiness, such as the increase of muscles and bone mass. Predicting testosterone levels can play an essential role in studying related diseases such as osteoporosis or prostate cancer that abnormalities of testosterone level might cause \citep{nieschlag2012testosterone}. We model the trend of testosterone levels across three basic body measures: age, waist circumference, and body mass index (BMI), motivated by clinical studies showing that aging and obesity are critical factors that reduce testosterone levels \citep[e.g.,][]{maarin1992effects}. In particular, we divide populations into three groups based on whether they smoke cigarettes every day (smokers), someday (occasional smokers), or do not (non-smokers). Some studies observed an interesting trend that people who smoke tend to have higher testosterone levels than those who do not smoke \citep[e.g.,][]{wang2013cigarette}. 

We collected data of male participants older than 25 sourced from NHANES in 2015-2016. In the NHANES dataset, the testosterone level is denoted as LBXTST. The attributes for age, waist circumference, and BMI are denoted as RIDAGEYR, BMXWAIST, and BMXBMI, respectively. We divided the participants into three groups based on the questionnaire categorizing participants' smoking habits into smokers, occasional smokers, and non-smokers. Here, we should note that the dataset includes observations for participants who did not describe their smoking habits, which constitute unlabeled observations. The number of observations for smokers, occasional smokers, non-smokers, and unreported is 372, 134, 674, and 980, respectively. We will call this dataset as \textit{Testosterone} dataset.

We perform two experiments. In the first experiment, we compare our model to the benchmark models similar to the insulin data application. We randomly chose 5\% of observations from each group and 100 observations from unlabeled observations to train the comparative models. We then evaluate trained models using the remaining labeled observations. The experiment will give us insights into the advantages of WSMGP compared to the benchmark models, specifically for the case of multidimensional inputs with more than two labels. In the second experiment, we compare WSMGP models trained with a different number of unlabeled observations. Specifically, to form a training dataset, we take 0, 50, 100, or 200 unlabeled observations while choosing labeled observations in the same way as the first experiment. We denote each case by WSMGP-$N^{u}$, e.g., WSMGP-0, WSMGP-50, and so on. Note that WSMGP-$0$ is equivalent to SCMGP. This experiment will shed light on the power of WSMGP leveraging unlabeled observations. All experiments are repeated 10 times with a standardized dataset. 

Table \ref{tab:tst} provides averages and standard deviations of RMSEs for the comparative models for each group. We see from the table that WSMGP predicts better than the benchmark models for all three groups. Again, the improvement is particularly significant for a sparse group (e.g., occasional smokers), as borrowing knowledge from dense groups in WSMGP through estimating between-output correlation can help infer an underlying function for the sparse group.  Also, the improved prediction of WSMGP upon SCMGP confirms the WSMGP's capability to exploit observations with missing labels. Moreover, SCMGP does not benefit from label estimation, observed from the deteriorated performance of SCMGP-LP. This is due to the significant overlap across outputs, which induces poor label estimation. Interestingly, contrary to the computation times per iteration, the total elapsed computation times of OMGP-WS and OMGP are greater than that of WSMGP. This indeed is unexpected because WSMGP estimates more parameters to consider potential between-group similarities. We conjecture that restricting the covariance structure to make estimated functions have a similar shape across groups in WSMGP might guide the model to find local minima easily, resulting in faster convergence.
\begin{table}[tbp]
  \centering
  \caption{Averages and standard deviations of RMSEs from 10 repeated experiments for the test data of Testosterone dataset. The best results are boldfaced. Computation times are in seconds. }
    \begin{tabular}{c|c c c c c}
    \hline
    \multirow{2}[0]{*}{Group} & \multicolumn{5}{c}{Average RMSE (standared deviation)} \\
\cline{2-6}          & WSMGP & OMGP-WS  & OMGP & SCMGP & SCMGP-LP \\
    \hline
    Smokers & \textbf{1.117} (0.16) & 1.152 (0.18) & 1.163 (0.14) & 1.206 (0.24) & 1.273 (0.21)\\
    Occasional smokers & \textbf{1.118} (0.19) & 1.184 (0.16) & 1.197 (0.20) & 1.255 (0.25) & 1.301 (0.17)\\
    Non-smokers & \textbf{0.898} (0.12) & 0.919 (0.18) & 0.919 (0.11) &  1.009 (0.30) & 0.996 (0.32)\\
    \hline
    Elapsed time (total) & 9.34 & 39.76 & 44.67 & 6.62 & 6.75\\
    Elapsed time (per iteration) & 0.3910 &	0.1251 &	0.1287 & 0.3029 & 0.3764\\
    \hline
    \end{tabular}%
  \label{tab:tst}%
\end{table}%

\begin{figure*}[h!]
\caption{RMSEs and computation times in different numbers of unlabeled observations.}
\vspace{0.2in}
\begin{center}
\centerline{\includegraphics[width=0.95\textwidth]{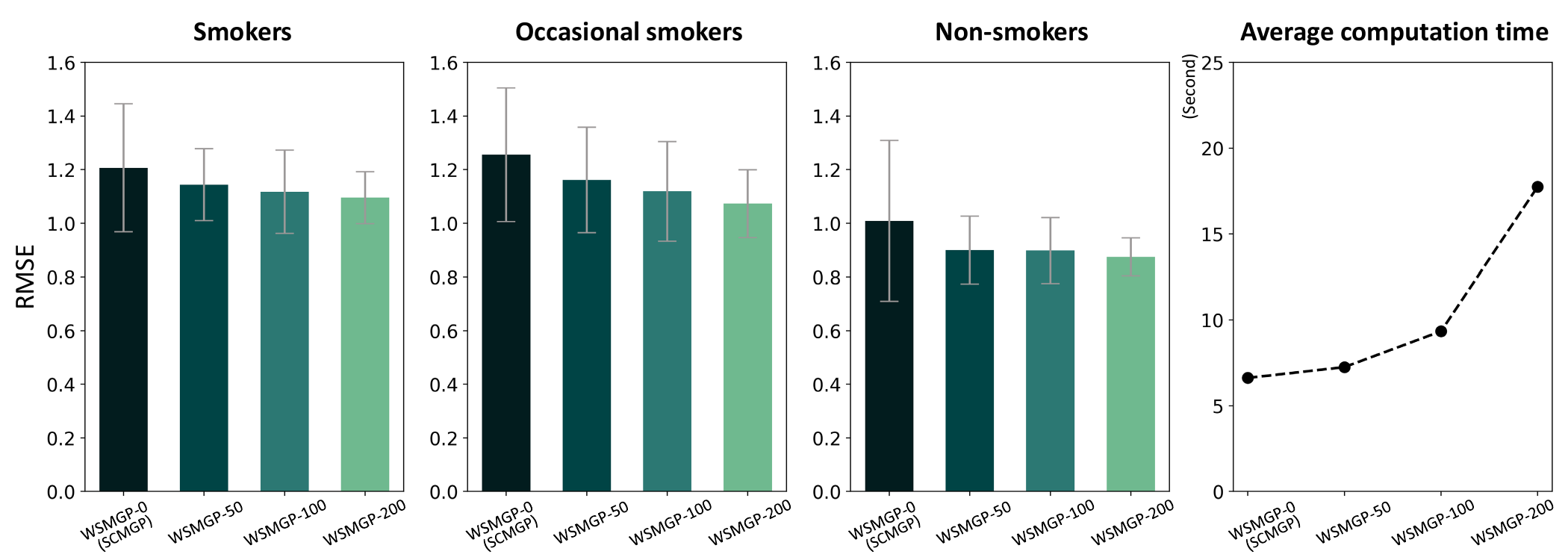}}
\label{fig:tst}
\end{center}
\vskip -0.3in
\end{figure*}

Fig. \ref{fig:tst} illustrates the trend of average RMSEs in the increasing unlabeled observations incorporated in the training dataset. We find that the prediction accuracy of WSMGP increases as more unlabeled observations are exploited, which implicates the necessity of using unlabeled data to improve predictions which WSMGP can attain. As we expected, incorporating more unlabeled observations results in an increase in computation time. Nonetheless, training and inference of WSMGP with a moderate number of observations is fast enough. Another intriguing observation is that the improvement in prediction becomes incremental as more unlabeled observations are used. It implies that including more observations will not be needed anymore once we have enough unlabeled observations. 

We finally recall that an additional case study on a bodyfat dataset can be found in Appendix \ref{sec:bodyfat}. We study a relationship between BMI and bodyfat percentage. Observations are collected from individuals grouped by gender, and we estimate a latent curve that relates measured BMIs to bodyfat percentages for each gender group. In particular, we examine our model on cases with different labeled fractions $(l)$ and the existence of missing ranges. We have seen consistently superior performance of our model compared to the benchmarks. For more detailed results, please refer to Appendix \ref{sec:bodyfat}.

\section{Conclusion \& Discussion}
\label{sec::conclusion}

This paper proposed a Bayesian probabilistic model that performs multi-output regression under a weakly-supervised setting. Our model is able to leverage (i) observations with missing group membership which is common in medical datasets (ii) prior belief on group membership, and (iii) correlation across multiple groups, in order to enhance prediction performance and uncover observation memberships. We highlight the advantageous properties of our approach through simulations and three real-world applications in healthcare.

Despite our focus on healthcare applications in the paper, the proposed model can be applied to different practices. Suppose that we seek to model individuals' yearly incomes based on several socio-demographic attributes (e.g., a place of residence, the number of households, or races) for different populations grouped by education levels. We expect that a potential between-group correlation should be estimated to obtain an accurate model for each group, which a multi-output regression model can address. However, we might have many people who do not disclose their education levels, that is, the observations with missing group labels. This example explains the necessity of developing a multi-output regression model capable of exploiting both unlabeled observations as well as an across-group correlation for better predictive accuracy. Indeed, many motivating examples may be derived in cases where group membership is missing or partially missing.

Our approach may also find application in both sequential decision-making and fairness. For instance, WSMGP can be an important ingredient in  the decision-making framework for sequential data acquisition. Our model is structured as a Bayesian hierarchical model, which naturally estimates the predictive uncertainty along with group membership probability for unlabeled observations. Such uncertainty information is extremely useful when deciding which unlabeled observations should be unveiled to achieve the best improvement in learning accuracy. This is also applicable to black-box optimization with noisy or missing labels \citep{xie2016bayesian, wang2020parallel} and online learning \citep{ryzhov2012knowledge}. Besides that, our case studies (insulin, testosterone, and bodyfat) shed light on improved fairness. Data from the minority group is often insufficient to build a good model and it is important to borrow strength across groups and learn from their commonalities. This however should be done in a manner that preserves unique features of data from minority groups. WSMGP is able to achieve this goal in our case studies.

Finally we note that WSMGP has some potential limitations. We build WSMGP based on the assumption that we have already determined the number of outputs ($M$). Unfortunately, we do not always recognize all possible groups if all labels from a certain group are missing (i.e., a ``hidden group''). The issue poses a crucial challenge in achieving fairness over groups because  WSMGP may not provide fair predictions if a separate curve for the hidden group is not inferred. This suggests an important extension of our method to automatically infer the number of groups $M$ from the data; thereby, the model can provide accurate predictions even for the hidden groups. Another possible challenge in WSMGP is the negative transfer of knowledge across groups when commonality is negligible \citep{kontar2020minimizing}. Although our model is flexible enough to predict each output separately (as each has its own latent function), negative transfer of knowledge may occur where the joint covariance may contaminate the prediction for an output by the knowledge from other uncorrelated outputs. Tackling the negative knowledge transfer issue under the weakly-supervised setting is a promising future direction. 

%
%
%

\newpage 
\begin{APPENDICES}
\section{Case Study on Bodyfat Dataset}\label{sec:bodyfat}

In this section, we utilize WSMGP for a case study using a \textit{Bodyfat} dataset, sourced from NHANES. 
The model settings for all comparative models are set to the same settings as in Sec. \ref{sec:experiment} in the original paper.    

The Bodyfat dataset relates BMI to bodyfat percentage measured from individuals. BMI is a value that is calculated by the weight divided by the square of the height of a person. The value has been used as a rough indicator of obesity \citep[e.g.,][]{decoda2008bmi}. Bodyfat percentage indicates the percentage of the total mass of fat to the total body mass. We collect the values of BMI and bodyfat percentage of individuals whose ages are above 19 from NHANES 2017-2018. Among different bodyfat percentages gauged by various methods, we use the measurements from Dual Energy X-ray Absorptiometry (denoted by DXDTOPF in the database). The total number of observations is 2240, grouped by gender with 1066 males and 1174 females. The curves for male and female groups that relate BMI to bodyfat percentage exhibit a clear correlation, as illustrated in Fig. \ref{fig:bodyfat-dataset}.

\begin{figure}[h!]
\caption{The Bodyfat dataset.}
\vspace{0.2in}
	\centering
	\includegraphics[width=0.5\textwidth]{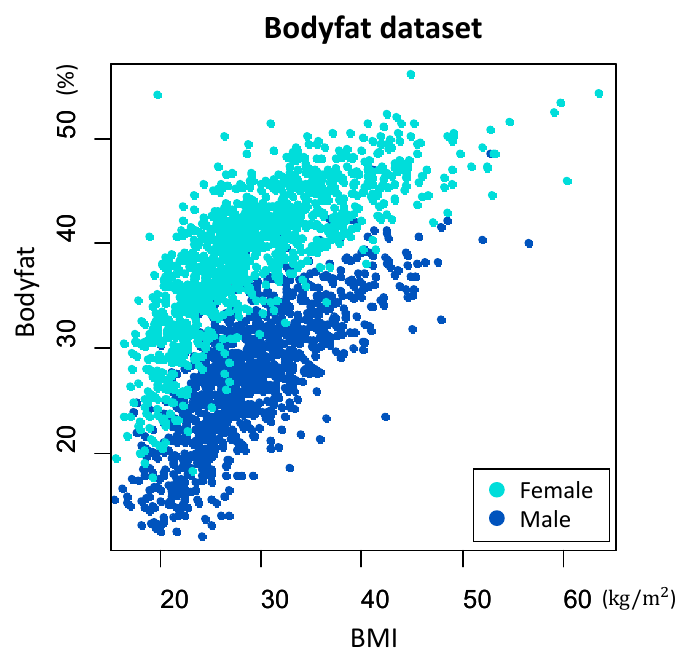}
	\label{fig:bodyfat-dataset}
\end{figure}

We randomly pick 224 observations (about 10\%) to establish our training data and set the remainder to the test data. For the training data, we randomly remove the gender information with three different labeled fractions: $l=0.1, 0.5,$ and $1$. We also consider two cases on missing ranges: one case is without any missing ranges, while the other case has missing ranges where the observations with the BMI in $(35, 40)$ and $(25, 30)$ for male and female are missing, respectively. That is, we perform the experiments on 6 cases in total. We design the case study to show the two key advantageous features of our model in a practical setting. First, with the different labeled fractions, we demonstrate our model's capability to efficiently infer both labels and predictions. Second, with the missing ranges, we highlight the benefits of accounting for correlation between groups. Note that the case with $l=0.1$ and missing ranges is the situation in which our model is expected to outperform the benchmarks, whereas the case with $l=1$ and no missing ranges is an adversarial setting to our model. For every case, we use RMSE as an evaluation quantity and repeat the experiment 10 times.  

{\begin{table}[t!]
 \centering
 \caption{Averages and standard deviations of RMSEs from 10 repeated experiments for the test data of Bodyfat dataset. The best result in each case is boldfaced.}
 \vspace{.3cm} 
    \adjustbox{max width=0.8\textwidth}{\begin{tabular}{c|c|c|cccc}
    \hline
    \multirow{2}[0]{*}{$l$} & \multirow{2}[0]{*}{Missing ranges exist?} & \multirow{2}[0]{*}{Group} & \multicolumn{4}{c}{Average RMSE (standard deviation)} \\
\cline{4-7}          &       &       & \textbf{WSMGP} & OMGP-WS & OMGP  & SCMGP \\
    \hline
    \multirow{4}[0]{*}{0.1} & \multirow{2}[0]{*}{No} & Female & \textbf{4.13} (0.05) & 4.23 (0.08) & 4.85 (0.90) & 6.32 (1.04) \\
         &       & Male  & \textbf{4.03} (0.12) & 4.08 (0.21) & 4.58 (0.83) & 6.72 (1.26) \\
\cline{2-7}          & \multirow{2}[0]{*}{Yes} & Female & \textbf{4.27} (0.16) & 4.60 (0.37) & 6.35 (0.50) & 5.47 (0.81) \\
         &       & Male  & \textbf{4.58} (0.79) & 5.66 (1.34) & 6.91 (1.35) & 6.30 (2.08) \\
    \hline
    \multirow{4}[0]{*}{0.5} & \multirow{2}[0]{*}{No} & Female & \textbf{4.12} (0.03) & 4.80 (0.95) & 4.86 (0.90) & 4.19 (0.13) \\
         &       & Male  & \textbf{3.95} (0.08) & 4.24 (0.83) & 4.65 (1.06) & 3.99 (0.10) \\
\cline{2-7}          & \multirow{2}[0]{*}{Yes} & Female & \textbf{4.18} (0.15) & 5.05 (0.98) & 6.32 (0.44) & 4.27 (0.21) \\
         &       & Male  & \textbf{4.04} (0.09) & 4.70 (1.01) & 7.40 (1.51) & 4.21 (0.22) \\
    \hline
    \multirow{4}[0]{*}{1} & \multirow{2}[0]{*}{No} & Female & \textbf{4.11} (0.02) & 4.37 (0.65) & 4.86 (0.91) & 4.11 (0.03) \\
         &       & Male  & 3.92 (0.07) & \textbf{3.92} (0.06) & 4.52 (0.87) & 3.92 (0.08) \\
\cline{2-7}          & \multirow{2}[0]{*}{Yes} & Female & \textbf{4.10} (0.05) & 4.38 (0.60) & 6.33 (0.42) & 4.12 (0.06) \\
         &       & Male  & \textbf{3.96} (0.08) & 3.99 (0.12) & 7.09 (1.40) & 3.99 (0.09) \\
    \hline
    \end{tabular}}%
 \label{tab:bodyfat-mse}%
\end{table}%
\begin{figure}[t!]
	\centering
	\includegraphics[width=0.9\textwidth]{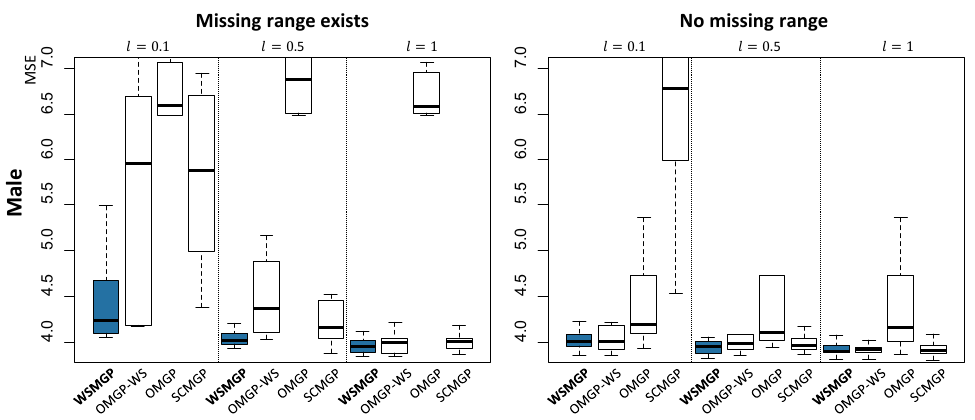}
	\caption{Boxplots of MSEs for Bodyfat data (male group).}\label{fig:bodyfat-boxplot}
\end{figure}
\begin{figure}[t!]
\caption{Illustration of the predicted curves with training data, predicted labels, and test data for Bodyfat dataset. The results are obtained under the settings $l=0.1$ with missing ranges. Note that WSMGP and OMGP-WS use both labeled (colored) and unlabeled (grey) data. OMGP uses the values of both labeled and unlabeled observations, but does not use label information. SCMGP uses labeled data only. Best viewed in color.}
\vspace{0.2in}
\begin{center}
\centerline{\includegraphics[width=0.9\textwidth]{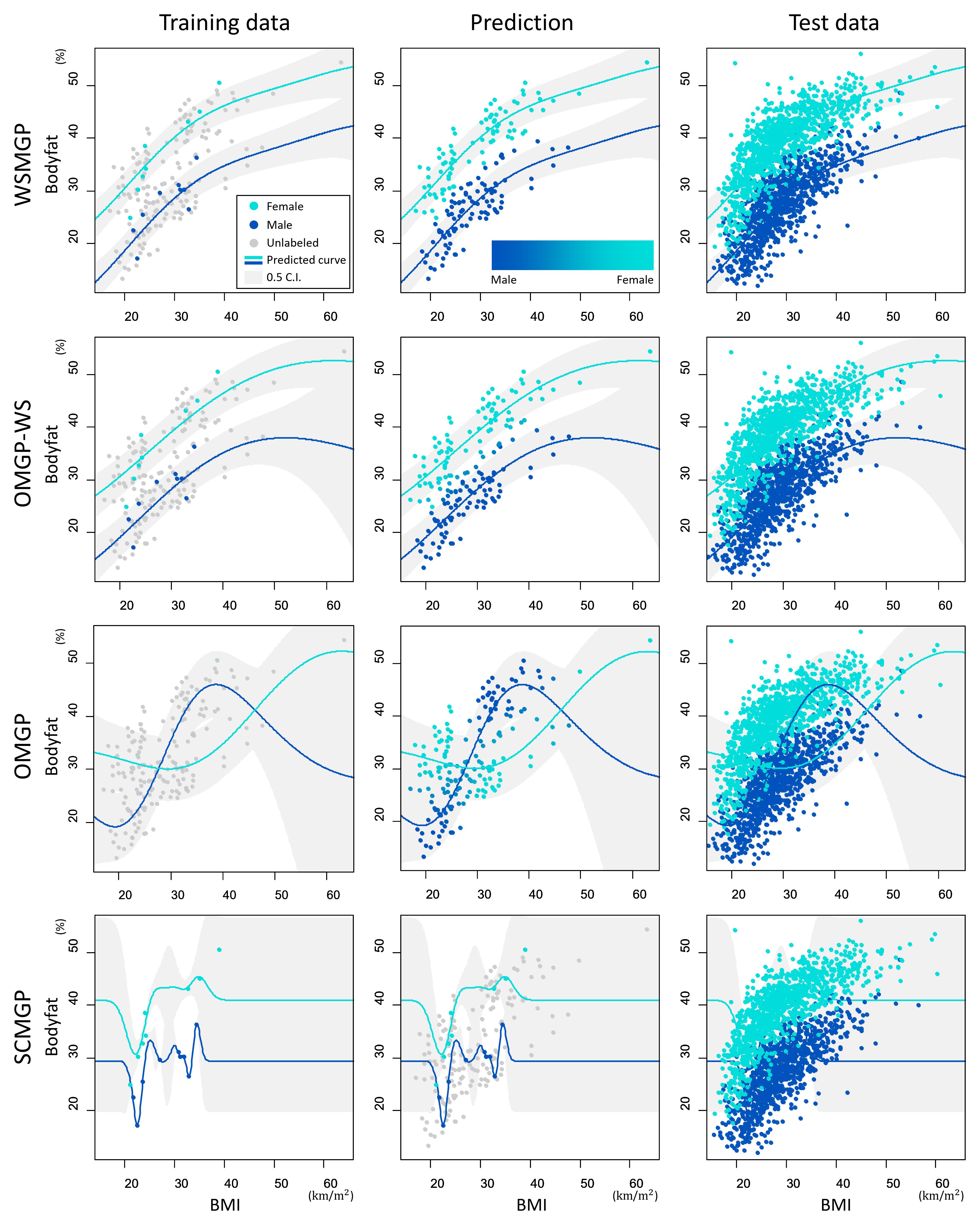}}
\label{fig:bodyfat-plots}
\end{center}
\end{figure}}

Experimental results are presented in Table \ref{tab:bodyfat-mse}, Fig. \ref{fig:bodyfat-boxplot} and Fig. \ref{fig:bodyfat-plots}. Table \ref{tab:bodyfat-mse} includes the RMSEs of the benchmark models for each case, where the results for males are summarized by Fig. \ref{fig:bodyfat-boxplot} in the form of boxplots. From the results, we find that WSMGP outperforms benchmark models for both female and male groups in most cases. Especially for the cases with $l=0.1$ and $0.5$, the prediction accuracy of WSMGP is clearly better than that of SCMGP which does not make use of unlabeled data. Fig. \ref{fig:bodyfat-plots} supports the result by showing the predicted curves in the case with $l=0.1$ and the missing ranges. We observe in the plots that WSMGP is able to effectively infer the underlying curves by using information from unlabeled data, whereas SCMGP fails to find the proper curves because the given labeled observations are not sufficient to characterize the general trends. Meanwhile, in the cases with missing ranges, WSMGP achieves smaller RMSEs compared to OMGP-WS and OMGP. This is because WSMGP accounts for the correlation between the two groups, whereas OMGP-WS and OMGP independently model the curves, often resulting in inaccurate predictions with high variance in the sparse regions. Note that the performance gap of WSMGP and OMGP-WS becomes smaller in the case without missing ranges and enough labels ($l=0.5$ or $1$), although WSMGP still outperforms OMGP-WS slightly. This is intuitively understandable as the given labels are sufficient for OMGP-WS to characterize the proper underlying curves even if they assume independence over groups. Another key insight, is that WSMGP attains competitive performances to OMGP-WS and SCMGP when observations are fully labeled ($l=1$) without missing ranges. In fact, this case corresponds to a traditional multi-output regression problem which is indeed an easier problem as accounting for correlation does not bring in substantially improved value. That being said, this result highlights the robustness of WSMGP to cases where sufficient labels/data are observed from all groups. Finally, we observe that as we have more labeled observations, the predictive accuracy of WSMGP, OMGP-WS, and SCMGP increases. Note that OMGP does not take advantage of the additional labels because they do not exploit the label information.

\section{Derivations of variational bounds}
 \label{appdx:der_vb}
 
 In this section, we provide detailed derivations of the two variational bounds, $\Lcvb$ and $\Lsvb$.
 
\subsection{Derivations of $\Lcvb$}

Here we use the same notations as in the main article. For notational simplicity, we omit hyperparameters $\{\boldsymbol{\theta}_{(\cdot)}, \boldsymbol{\alpha}_0\}$ in the derivations unless it causes confusion. We use the notation $\mathbb{E}_{\mathbf{a}}[ \cdot ]$ to indicate expectation over a variational distribution $q(\mathbf{a})$.


First, we start with calculating the following expectation that will be used shortly:
\begin{equation}\label{eq:a1}
\begin{split}
   \mathbb{E}_{\mathbf{Z}}[\log p(\mathbf{y} | \mathbf{f}, \mathbf{Z})] &= \mathbb{E}_{\mathbf{Z}}\left[\sum_{n=1, m=1}^{N,M}[\mathbf{Z}]_{nm} \log \mathcal{N}(\mathbf{y}_n |[\mathbf{f}_m]_n, \sigma_m^2)\right] \\
&= \sum_{n=1, m=1}^{N,M} [\hat{\mathbf{\Pi}}]_{nm} \log \mathcal{N}(\mathbf{y}_n | [\mathbf{f}_m]_n, \sigma_m^2) \\
& = \sum_{n=1, m=1}^{N,M} \log \mathcal{N}(\mathbf{y}_n | [\mathbf{f}_m]_n, \frac{\sigma_m^2}{[\hat{\mathbf{\Pi}}]_{nm}}) + \frac{1}{2} \sum_{n=1,m=1}^{N,M}\log\frac{(2\pi\sigma_m^2)^{(1-[\hat{\mathbf{\Pi}}]_{nm})}}{[\hat{\mathbf{\Pi}}]_{nm}}.\\
& = \log \N (\y|\F, \D) + \frac{1}{2} \sum_{n=1,m=1}^{N,M}\log\frac{(2\pi\sigma_m^2)^{(1-[\hat{\mathbf{\Pi}}]_{nm})}}{[\hat{\mathbf{\Pi}}]_{nm}} 
\end{split} 
\end{equation}

Now we derive $\Lcvb$. Recall that we have 
\begin{equation}\label{eq:a3}
\begin{split}
&\Lcvb = \log \int_{\F, \mathbf{u}} \bigg( e^{\int_{\Z,\bPi^u} q(\Z)q(\bPi^u) \log \frac{ p(\y | \F, \Z) p(\Z | \bPi) p(\bPi^u)}{q(\Z)q(\bPi^u)}} p(\F | \mathbf{u}, \X, \W) p(\mathbf{u}| \W)\bigg)\leq \log p(\y|\X,\W)
. 
\end{split}
\end{equation}
In Eq. \eqref{eq:a3}, we first focus on the following integral
\begin{equation}\label{eq:a4}
\begin{split}
&\int_{\Z,\bPi^u} q(\Z)q(\bPi^u) \log \frac{ p(\y | \F, \Z) p(\Z | \bPi) p(\bPi^u)}{q(\Z)q(\bPi^u)} \\
&\quad= \mathbb{E}_{\mathbf{Z}}[\log p(\mathbf{y} | \mathbf{f}, \mathbf{Z})] -\KL(q(\Z^l)\Vert p(\Z^l | \bPi^l)) -\KL(q(\Z^u)q(\bPi^u)||p(\Z^u | \bPi^u)p(\bPi^u))= \log \N (\y|\F, \D) + \mathcal{V} 
\end{split}
\end{equation}
where the last identity is based on Eq. \eqref{eq:a1} and the terms for $\Z, \bPi$ are collected into
\begin{equation*}\label{eq:a4-1}
\begin{split} 
\mathcal{V} &= \frac{1}{2} \sum_{n=1,m=1}^{N,M}\log\frac{(2\pi\sigma_m^2)^{(1-[\hat{\mathbf{\Pi}}]_{nm})}}{[\hat{\mathbf{\Pi}}]_{nm}}-\KL(q(\Z^l)||p(\Z^l | \bPi^l)) -\KL(q(\Z^u)q(\bPi^u)||p(\Z^u | \bPi^u)p(\bPi^u)). 
\end{split}
\end{equation*}
Plugging Eq. \eqref{eq:a4} into Eq. \eqref{eq:a3}, we obtain the final form of $\Lcvb$ as
\begin{equation}\label{eq:a5}
\begin{split}
	\Lcvb & = \log \int_{\F, \mathbf{u}} \bigg( \exp{\int_{\Z,\bPi^u} q(\Z)q(\bPi^u) \log \frac{ p(\y | \F, \Z) p(\Z | \bPi) p(\bPi^u)}{q(\Z)q(\bPi^u)}}p(\F | \mathbf{u}, \X, \W) p(\mathbf{u}| \W)\bigg)\\
	&= \log \int_{\F, \mathbf{u}}   \N (\y|\F, \D) p(\mathbf{f}|\mathbf{u}, \mathbf{X}, \mathbf{W})p(\mathbf{u}| \mathbf{W}) + \mathcal{V}\\
	&= \log \N(\y | \zero, \B + \KFu\K^{-1}_{\mathbf{u},\mathbf{u}}\KuF + \D) + \mathcal{V}
\end{split}
\end{equation}
of which the last equality is based on Eq. \eqref{eq:scmgp} in the main article.

\subsection{Derivations of $\Lsvb$}
To obtain the scalable variational bound $\Lsvb$, we further introduce $q(\bu) = \mathcal{N}(\bmu_\bu, \bS)$ to approximate $p(\bu)$. We first derive the variational marginalized distribution for $\Fm$ as 
\begin{equation*}
\begin{split}
    q(\Fm) & = \int_\bu q(\Fm, \bu) = \int_\bu p(\Fm \vert \bu,\X, \W) q(\bu) = \mathcal{N}\left( \K_{\Fm,\mathbf{u}}\Kuu\boldsymbol{\mu}_\bu,
\K_{\Fm,\Fm} + \K_{\Fm,\mathbf{u}} \Kuuinv(\mathbf{S} - \Kuu)\Kuuinv\K_{\mathbf{u},\Fm}\right).
\end{split}
\end{equation*}
Then, the lower bound $\Lsvb$ is given by 
\begin{equation*}
\begin{split}
    \Lsvb &= \int_{\F, \bu, \Z, \bPi} q(\F, \bu) q(\Z) q(\bPi^u) \log \frac{p(\Y \vert \F, \Z) p(\F | \mathbf{u}, \X, \W) p(\mathbf{u}| \W)p(\Z | \bPi) p(\bPi^u)}{q(\F, \bu) q(\Z) q(\bPi^u) }\\
    &=  \mathbb{E}_{\mathbf{\F}} \left[ \mathbb{E}_{\mathbf{Z}}[\log p(\mathbf{y} | \mathbf{f}, \mathbf{Z})] \right] -\KL(q(\bu)\Vert p(\bu))-\KL(q(\Z^l)\Vert p(\Z^l | \bPi^l)-\KL(q(\Z^u)q(\bPi^u))\Vert p(\Z^u | \bPi^u)p(\bPi^u))\\
    &= \mathbb{E}_{\mathbf{\F}}\left[ \log \mathcal{N}(\Y \vert \F, \D)   \right] -\KL(q(\bu)\Vert p(\bu)) + \mathcal{V},
\end{split}
\end{equation*}
where we use the variational expectation Eq. \eqref{eq:a1} for the last identity. We recover Eq. \eqref{svb} by noting that $\log \mathcal{N}(\Y \vert \F, \D) = \sum_{n=1,m=1}^{N,M} \log \mathcal{N}([\Y]_n \vert [\Fm]_n, \frac{\sigma^2_m}{[\hat \bPi]_{nm}})$ because $\D$ is a diagonal matrix. 

\section{Gradients of Variational Bounds}
\label{appdx:der_grad}
To utilize gradient-based optimization algorithms, we provide the gradients of proposed variational bounds for the parameters.  In $\Lcvb$, the parameters to be optimized are: $\btheta$, $\bsigma$, $\alpha_0$, $\hat \bPi$ and $\W$, where $\btheta$ collects the hyperparameters related to $p(\F \vert \X)$ and $p(\bu | \W)$, denoted by $\btheta_\F$ and $\btheta_\bu$, respectively. In $\Lsvb$, we additionally have $\bmu_\bu$ and $\bS$. 

We first consider $\mathcal{V}$ that appears in both $\Lcvb$ and $\Lsvb$. The related parameters are $\bbalpha_0$ and $\hat \bPi$, that is, $\alpha_0$ and $[\hat \bPi]_{nm}$ for $n=1,...,N$ and $m=1,...,M$. Because $\mathcal{V}$ can be expressed as the summation in which terms are independent of each other in terms of the parameters, it is trivial to derive the partial derivatives; we omit the derivatives of $\mathcal{V}$.  

The remaining terms are a bit tricky to obtain partial derivatives since they include matrices. In order to obtain the partial derivatives for the parameters in matrice, we use the notation in \cite{brookes2020matrix}: we define $\bG \mvec$, which denotes a vector obtained by vectorizing the matrix $\bG$. The partial derivative for each parameter is calculated using the law of total derivative and the chain rule. For example, consider $\btheta_\F$ in $\Lcvb$. The matrices that involve $\btheta_\F$ are $\KFu$ and $\KFF$.  Hence, the partial derivative of $\btheta_\F$ is given by $\frac{\partial\Lcvb}{\partial\btheta_\F} = \frac{\partial\Lcvb}{\partial\KFu\mvec}\frac{\partial\KFu\mvec}{\partial\btheta_\F} + \frac{\partial\Lcvb}{\partial\KFF\mvec}\frac{\partial\KFF\mvec}{\partial\btheta_\F}.$ In the following sections, we find the partial derivatives of the proposed variational bounds for each matrix.

\subsection{Matrix Derivatives of $\Lcvb$}
 
Let $\Lcvbone$ denote the first term in $\Lcvb$, $\log \N(\y | \zero, \B + \KFu\K^{-1}_{\mathbf{u},\mathbf{u}}\KuF + \D)$. Note that this is structurally equivalent to the marginal distribution of sparse convolved multi-output GP (1) proposed by \citet{alvarez2011computationally}. Our derivation is similar to their work, we thus directly provide the results. The reader is referred to the supplement of \citet{alvarez2011computationally} for detailed derivations.
\begin{equation*}
\begin{split}
    \frac{\partial\Lcvbone}{\partial\KFF\mvec} = \frac{\partial\Lcvbone}{\partial\D\mvec} &= -\frac{1}{2} \left(\bQ\mvec \right)^\top,\\
    \frac{\partial\Lcvbone}{\partial\KuF\mvec} &= \left(\left( \Kuuinv \KuF \bQ  - \bC \KuF \Binv + \Ainv \KuF \Binv \Y \Y^\top \Binv \ \right)\mvec\right)^\top,\\
    \frac{\partial\Lcvbone}{\partial\Kuu\mvec} &= -\frac{1}{2}\left(\left(\Kuuinv - \bC - \Kuuinv \KuF \bQ \KFu \Kuuinv \right)\mvec\right)^\top,
\end{split}
\end{equation*}
with $\bQ = \left( \Binv \bJ \Binv \odot \bP \right)$; $\bJ = \B - \Y \Y^\top + \KFu \bA^{-1} \KuF \B^{-1} \Y \Y^\top + \left( \KFu \bA^{-1} \KuF \B^{-1} \Y \Y^\top \right)^\top - \KFu \bC \KuF$; $\bC = \Ainv + \Ainv\KuF\Binv\Y\Y^\top\Binv\KFu\Ainv$, where $\odot$ is the Hadamard product and $\bP = \I_M \otimes \mathbf{1}_{N \times N}$, where $\mathbf{1}_{N \times N}$ denotes the $N \times N$ matrix with ones and $\otimes$ is the Kronecker product.

\subsection{Matrix Derivatives of $\Lsvb$}
For notational simplicity, let us define $\Lsvbone = \mathbb{E}_{\mathbf{\F}}\left[ \log \mathcal{N}(\Y \vert \F, \D)\right]$ and $\Lsvbtwo = \KL(q(\bu)\Vert p(\bu))$, which are the first and second term in $\Lsvb$, respectively. That is, $\Lsvb = \Lsvbone - \Lsvbtwo + \mathcal{V}$. We need to find (i) the matrix partial derivatives for $\Lsvbone$:  $\frac{\partial\Lsvbone}{\partial\bmu_\bu}, \frac{\partial\Lsvbone}{\partial \bS \mvec}, \frac{\partial\Lsvbone}{\partial \Kuu \mvec},  $ and (ii) the matrix partial derivatives for $\Lsvbtwo$: $ \frac{\partial\Lsvbtwo}{\partial \bmu_\bu}, \frac{\partial\Lsvbtwo}{\partial \bS\mvec}, \frac{\partial\Lsvbtwo}{\partial \Kuu\mvec}, \frac{\partial\Lsvbtwo}{\partial \KuF\mvec},$ $
\frac{\partial\Lsvbtwo}{\partial \textrm{diag}(\KFF)},
\frac{\partial\Lsvbtwo}{\partial \textrm{diag}(\D)}$. To do this, we are first required to find the derivatives of the variational expectation
\begin{equation*}
    \begin{split}
        \frac{\partial}{\partial\bmu_{q(\fm)}} \mathbb{E}_{\Fm}\left[\log \mathcal{N}(\Y_m \vert \Fm, \D_m))\right] &= (\Y_m - \bmu_{q(\fm)}) \odot \textrm{diag}(\D_m), \\
        \frac{\partial}{\partial\bSigma_{q(\fm)}} \mathbb{E}_{\Fm}\left[\log \mathcal{N}(\Y_m \vert \Fm, \D_m))\right] &= -\frac{1}{2}  \textrm{diag}(\tilde\D_m),
    \end{split}
\end{equation*}
where $\tilde\D_m$ represents the Hadamard reciprocal of $\D_m$. Setting $\frac{\partial}{\partial\bmu_{q(\fm)}} \mathbb{E}_{\Fm}\left[\log \mathcal{N}(\Y_m \vert \Fm, \D_m))\right] = \ba_m$ and $\frac{\partial}{\partial\bSigma_{q(\fm)}} \mathbb{E}_{\Fm}\left[\log \mathcal{N}(\Y_m \vert \Fm, \D_m))\right] = \bv_m$, then the partial matrix derivatives for $\Lsvbone$ and $\Lsvbtwo$ are derived as 

\begingroup
\allowdisplaybreaks
\begin{align*}
        \frac{\partial\Lsvbone}{\partial\bmu_\bu} &= \sum_{m=1}^{M} \left( \Kuuinv\KuFm\ba_m  \right)\\
        \frac{\partial\Lsvbone}{\partial\bS\mvec} &= \left(\left(\sum_{m=1}^{M} \bH_m \right)\mvec\right)^\top\\
        \frac{\partial\Lsvbone}{\partial\Kuu\mvec} &=\frac{1}{2}\left(\left(\bR + \bR^\top\right)\mvec\right)^\top \\
        \frac{\partial\Lsvbone}{\partial\KuFm\mvec} &= \left(\left( \Kuuinv \bmu_\bu \ba_m^\top +  2\left(\bS \Kuuinv - \I_{Q\times Q}\right)^\top \bO_m\right)\mvec\right)^\top\\
        \frac{\partial\Lsvbone}{\partial\textrm{diag}(\Kfmfm)} &= \bv_m \\
        \frac{\partial\Lsvbone}{\partial\textrm{diag}(\D_m)} &= \frac{1}{2} \left( \bv_m + (\y_m \odot \y_m + \bmu_\bu \odot \bmu_\bu + \textrm{diag}(\bSigma_{q(\Fm)}) - 2\bmu_\bu \odot \Y_m)\odot \textrm{diag}(\D_m \odot\D_m)  \right) 
\end{align*}
\endgroup
and
\begin{equation*}
    \begin{split}
        \frac{\partial\Lsvbtwo}{\partial\bmu_\bu} &= \Kuuinv \bmu_\bu,\\
        \frac{\partial\Lsvbtwo}{\partial\bS\mvec} &= \frac{1}{2}\left(\left(\Kuuinv - \bS \right)\mvec\right)^\top \\
        \frac{\partial\Lsvbtwo}{\partial\Kuu\mvec} &= \frac{1}{2}\left(\left(- (\Kuuinv \bS \Kuuinv)^\top - (\Kuuinv)^\top \bmu_\bu\bmu_\bu^\top(\Kuuinv)^\top + (\Kuuinv)^\top  \right)\mvec\right)^\top,
    \end{split}
\end{equation*}
where $\bH_m = \bO_m \KFmu \Kuuinv $; $\bO_m = \Kuuinv\KuFm \odot \left( \mathbf{1}_{Q\times1} \otimes \bv_m^\top \right)$; $\bR = \sum_{m=1}^{M}\bigg(\bH_m - \bH_m \bS \Kuuinv - \left(\bH_m \bS \Kuuinv\right)^\top - \Kuuinv \KuFm \ba_m \left( \Kuuinv \bmu_\bu \right)^\top\bigg)$. 

\end{APPENDICES}



\bibliographystyle{informs2014} 



\bibliography{ref}

\end{document}